\documentclass[lettersize,journal]{IEEEtran}
\usepackage{amsmath,amsfonts}
\usepackage{algorithmic}
\usepackage{algorithm}
\usepackage{array}
\usepackage[
    caption=false,
    font=footnotesize,
    labelfont=rm,
    textfont=rm
]{subfig}
\usepackage{textcomp}
\usepackage{stfloats}
\usepackage{url}
\usepackage{verbatim}
\usepackage{graphicx}
\usepackage{cite}
\usepackage{booktabs,tabularx}
\newcommand{\IIRTableStyle}{\footnotesize\renewcommand{\arraystretch}{1.16}\setlength{\tabcolsep}{3pt}}
\usepackage{orcidlink}

\begin{document}

\title{Radar-Based 3D Human Pose Estimation via Motion-Aware Pose Initialization and Observation-Guided Refinement}

\author{
Woo-Jin Jung, 
Chan-Bin Lim, 
and 
Seung-Hyun Kong*,
\thanks{This work was supported by the Ministry of SMEs and Startups (MSS) grant funded by the Korean government (No. RS-2025-24535910). (Corresponding author: Seung-Hyun Kong.)}
\thanks{Woo-Jin Jung and Seung-Hyun Kong are with the CCS Graduate School of Mobility, Korea Advanced Institute of Science and Technology, Daejeon 34051, South Korea (e-mail: woo-jin.jung@kaist.ac.kr; skong@kaist.ac.kr).}
\thanks{Chan-Bin Lim is with the Robotics Program, Korea Advanced Institute of Science and Technology, Daejeon 34141, South Korea (e-mail: chanbin.lim@kaist.ac.kr).}
}



\maketitle

\begin{abstract}
3D Human Pose Estimation (3D HPE) is used in indoor environments, such as smart homes, to analyze human movement and monitor human states. Millimeter-wave (mmWave) radar is suitable for this task because it is robust to lighting variations and does not capture images of human appearance. However, radar observations often provide insufficient information about certain body parts or contain overlapping reflections from multiple body parts, limiting accurate 3D joint localization. To address these limitations, we propose the Initialization-to-Refinement Pose Framework (I2R-Pose), comprising two stages: Motion-Aware Pose Initialization and Observation-Guided Refinement. In the first stage, joint-wise features are learned by incorporating temporal relationships across consecutive radar observations and human skeletal structure to compensate for insufficient joint information and estimate an initial 3D pose sequence. In the second stage, separately extracted radar observation features predict joint-wise positional corrections that refine the initial pose and generate the final pose sequence. The framework supports individual and fused inputs of time–Doppler/time–range (TD/TR) maps and radar point clouds (PCs). To evaluate performance across indoor activities, we construct the Radar Motion and Spatial Information for 3D Human Pose Estimation (RaMoS-Pose) dataset, providing synchronized TD/TR maps, radar PCs, and 3D joint ground truth (GT). Evaluations on RaMoS-Pose and the publicly available MVDoppler-Pose dataset show that I2R-Pose reduces mean per-joint position error (MPJPE) by 39.50\% on MVDoppler-Pose and by 15.13\% on RaMoS-Pose with fused inputs, compared with the best-performing existing methods evaluated on each dataset.
\end{abstract}



\section{Introduction}
3D Human Pose Estimation (3D HPE) aims to estimate the 3D positions of human body joints from sensor observations and is widely used in various applications, including motion analysis, rehabilitation monitoring, and human--machine interaction~\cite{an2022health_mri,roshandel2025mmprivpose3d,zhao2018rf3dskeletons}.
In Internet of Things (IoT)-enabled indoor environments, such as smart homes, non-contact human pose estimation requires consideration of both lighting variations and privacy concerns associated with image acquisition~\cite{zhang2023mmwavesurvey,cao2025occmesh}.
From this perspective, millimeter-wave (mmWave) radar has attracted attention as an alternative to camera-based HPE because it is less dependent on lighting conditions and does not directly capture images of human appearance~\cite{sengupta2020mmpose,multiscale_hpe,shi2025mmpointattention}.

In radar-based HPE, radar point clouds (PCs), which provide the 3D positions of reflection points on the human body, have been widely used~\cite{xue2021mmmesh,cao2024virteach}.
However, radar PCs may lack sufficient spatial information for estimating joint positions because reflection points from certain body parts are not adequately detected.
This is because radar reflection characteristics vary across body parts, and the number and distribution of detected reflection points change depending on body orientation and posture~\cite{radar_specularity,wang2023human,gao2026stargraph}.
To compensate for these insufficient observations, previous studies have utilized spatiotemporal features from multiple frames, skeletal relationships between joints, and point-wise Doppler information to improve joint position estimation for body parts with sparse reflection points~\cite{sengupta2023mmposenlp,hu2024mmposefk,multiscale_hpe,vcf-pose}.

\begin{figure}[t]
    \centering
    \includegraphics[width=\columnwidth]
    {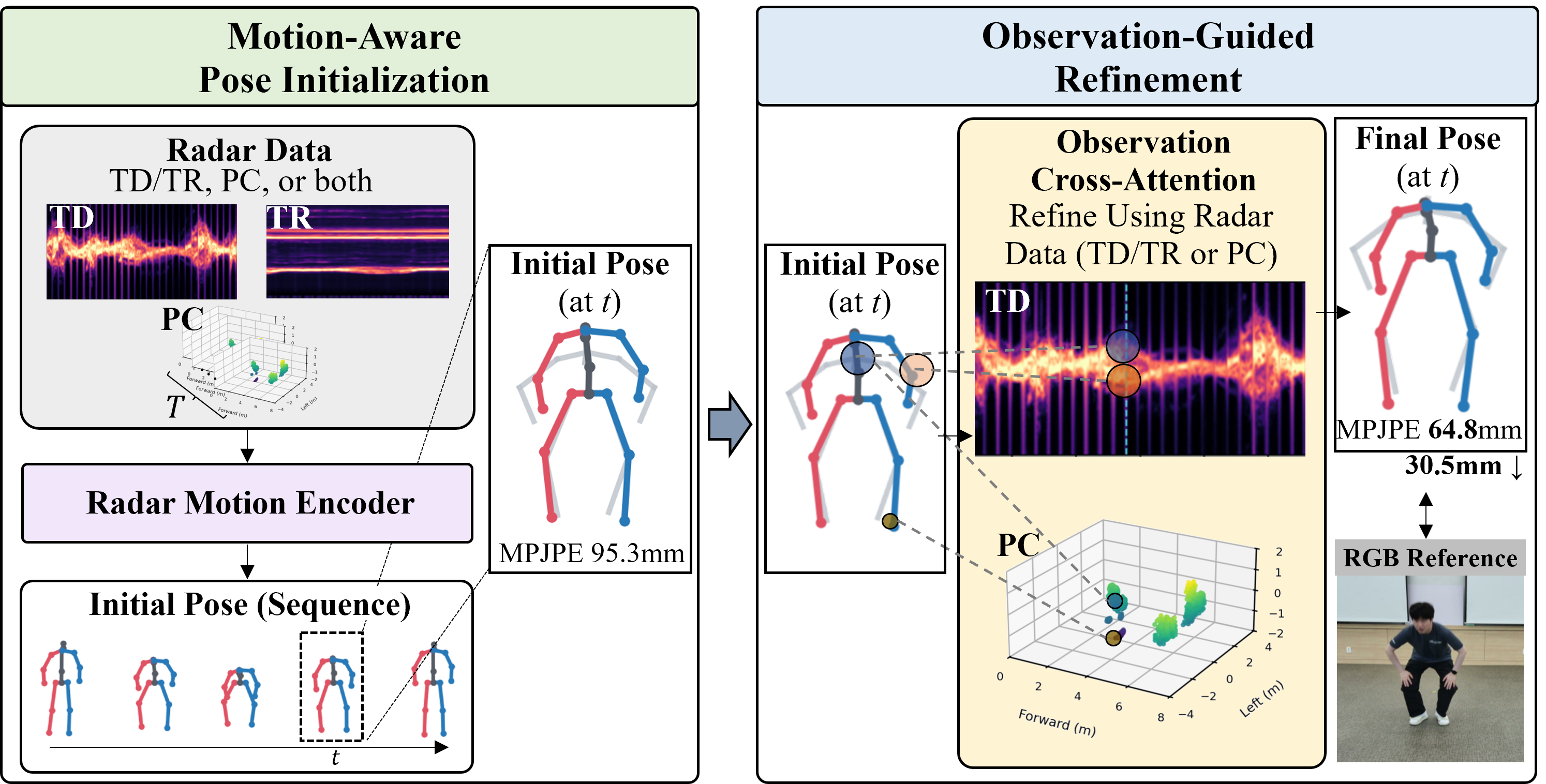}
    \caption{Conceptual overview of I2R-Pose. In the Motion-Aware Pose Initialization stage, joint-wise features are learned by incorporating temporal relationships across consecutive radar observations (TD/TR maps, PCs, or both) and human skeletal structure to estimate an initial 3D pose sequence. In the subsequent Observation-Guided Refinement stage, radar observation features extracted separately from the same radar input are combined with the joint-wise features used for initial pose estimation to predict joint-wise positional corrections. These corrections are applied to the initial pose to generate the final pose sequence. The refinement process and changes in MPJPE are illustrated using a representative frame, while RGB images are provided solely for visual reference.}
    \label{fig:overview}
\end{figure}

In addition to methods using radar PCs, human pose estimation has also been investigated using temporal variations in radar reflection signals.
Radar provides motion information through Doppler frequency shifts caused by the radial relative velocities of reflecting objects~\cite{chen2006micro,zhang2024radarfeatures, song2026star}.
The \textit{time--Doppler (TD) map}, which represents the Doppler distribution over time, has been used for human activity recognition and pose estimation~\cite{kim2009human_microdoppler,zhou2023mdpose,tang2024mdpose}.
Recently, a method combining TD maps with \textit{time--range (TR) maps}, which represent the distribution of reflected signals across ranges over time, has been proposed to jointly utilize motion and range information~\cite{mvdoppler_pose}.
However, reflection signals from multiple body parts may overlap in TD/TR maps.
This overlap makes it difficult to distinguish which body parts contribute to the observed Doppler and range information, thereby limiting the accurate estimation of individual joint positions in 3D space~\cite{qiao2022mdseparation,ren2018stssm}.

As described above, radar PCs may provide insufficient spatial information for certain body parts, while TD/TR maps may contain overlapping reflection signals from multiple body parts, making accurate estimation of individual joint positions in 3D space challenging.
However, human joint movements are temporally continuous and correlated through skeletal structure, and these relationships can provide useful cues for estimating joint positions of body parts with insufficient observations~\cite{pavllo20193d,rempe2021humor}.
Nevertheless, even when these relationships are utilized for pose estimation, discrepancies may remain between the estimated and actual joint positions at individual time steps.
Therefore, a sequential approach can be considered in which an initial pose is first estimated and then refined at the joint level using separately extracted radar observation features.

Based on this approach, this paper proposes the \textit{Initialization-to-Refinement Pose Framework (I2R-Pose)}, a two-stage framework that sequentially performs initial pose estimation and radar observation-based refinement, as illustrated in Fig.~\ref{fig:overview}.
First, in the \textit{Motion-Aware Pose Initialization} stage, the Radar Motion Encoder learns joint-wise features by incorporating temporal relationships across consecutive radar observations and human skeletal structure, and these features are used to estimate an initial 3D pose sequence.
During training, motion representation learning is further guided by aligning radar-based features with motion features extracted from ground-truth (GT) joint sequences.
Subsequently, in the \textit{Observation-Guided Refinement} stage, a separate Radar Observation Encoder extracts observation features from the same radar input to predict joint-wise positional corrections that reduce residual joint localization errors in the initial pose.
The predicted corrections are added to the initial pose to generate the final 3D pose sequence.
I2R-Pose is not limited to a specific radar observation representation and supports individual inputs of TD/TR maps and radar PCs by adapting the architectures of the Radar Motion Encoder and Radar Observation Encoder to the input characteristics.
When both types of observations are available, they are fused through joint-wise features for pose estimation. 

The proposed method is evaluated on the publicly available MVDoppler-Pose dataset~\cite{mvdoppler_pose} and the \textit{Radar Motion and Spatial Information for 3D Human Pose Estimation (RaMoS-Pose)} dataset constructed in this study.
RaMoS-Pose provides synchronized TD/TR maps, radar PCs, and 3D joint GT for various indoor postures and activities, enabling the evaluation of pose estimation performance under different input modalities and fusion configurations.
On MVDoppler-Pose, I2R-Pose achieves an MPJPE of 36.692~mm, representing a 39.50\% reduction compared with existing models using TD/TR maps.
On RaMoS-Pose, the PC-only configuration achieves an MPJPE of 129.311~mm, reducing the error by 14.42\% compared with the best-performing existing PC-based method.
Furthermore, the TD/TR--PC fusion configuration achieves an MPJPE of 128.237~mm, yielding the lowest error among the evaluated configurations.

The main contributions of this study are summarized as follows:

\begin{itemize}

    \item \textbf{Motion-aware initial pose estimation to compensate for insufficient joint information:}
    We propose a method for estimating an initial 3D pose sequence by learning joint-wise features that incorporate temporal relationships across consecutive radar observations and human skeletal structure.
    During training, motion representation learning is guided by aligning radar-based features with GT-derived motion features.
    When both TD/TR maps and PCs are available, the two types of observations are fused through joint-wise features.

    \item \textbf{Observation-guided pose refinement to reduce joint localization errors:}
    We propose a refinement method that utilizes separately extracted radar observation features to reduce residual joint localization errors in the initial pose.
    Joint-wise positional corrections are predicted by combining the joint-wise features used for initial pose estimation with radar observation features, and these corrections are applied to the initial pose to generate the final 3D pose.

    \item \textbf{Construction and evaluation of a dataset covering diverse indoor activities:}
    We construct the RaMoS-Pose dataset, which provides synchronized TD/TR maps, radar PCs, and 3D joint GT.
    We compare the proposed method with existing approaches on both public and newly constructed datasets and evaluate pose estimation performance across major model components and radar input configurations.

\end{itemize}


\section{Related Work}
\label{sec:related_works}

This section reviews prior studies related to I2R-Pose, focusing on radar-based HPE methods and datasets.
First, radar-based HPE methods are compared by categorizing them into point-cloud-based and time--Doppler-based approaches, followed by an overview of the characteristics of existing radar HPE datasets.

\subsection{Radar-Based Human Pose Estimation Methods}
\label{subsec:radar_hpe_methods}

\subsubsection{Point-Cloud-Based Methods}
\label{subsec:related_PC_based}

Radar point-cloud-based HPE methods have evolved to estimate human joint positions using the spatial information of irregularly distributed reflection points.
mm-Pose~\cite{sengupta2020mmpose} converted reflection point information into an image representation and estimated joint positions using a convolutional neural network (CNN).
mmPose-NLP~\cite{sengupta2023mmposenlp} organized point clouds from multiple frames into voxel sequences and utilized temporal context through sequence-to-sequence learning.
mmParse~\cite{wang2023human} jointly learned point-level semantic information and human body structure through PointNet-based feature extraction and joint training of body-part segmentation and pose estimation.

To compensate for the sparsity and irregularity of radar observations, studies have also explored the use of human body structure and spatiotemporal features.
mmPose-FK~\cite{hu2024mmposefk} incorporated forward kinematics into a neural network to impose structural constraints of the human body.
mmHPE~\cite{wu2025mmhpe} combined point cloud generation based on detected human regions with multi-scale coarse-to-fine pose estimation.
MS-STPoseNet~\cite{multiscale_hpe} learned hierarchical spatial features and multi-scale temporal features to jointly model human spatial structure and temporal motion.
VCF-Pose~\cite{vcf-pose} estimated an initial pose from sequential PCs and subsequently refined it using Doppler-based velocity features and human structural information.

These studies have utilized not only the spatial information of radar PCs but also temporal context and human body structure, with some adopting a sequential refinement of the initial pose.
In contrast, our study estimates an initial pose by learning joint-wise features that incorporate motion and skeletal relationships from consecutive radar observations, assisted by GT-derived motion feature alignment during training.
Subsequently, joint positions are refined using separately extracted radar observation features.

\subsubsection{Time--Doppler-Based Methods}
\label{subsec:related_TD_based}

Time--Doppler (TD) maps represent changes in the Doppler distribution over time and provide features related to human motion.
Doppler information has been utilized for human activity recognition~\cite{kim2009human_microdoppler}, and methods extending its use to human pose estimation have also been proposed.
MD-Pose~\cite{zhou2023mdpose} applied a U-Net-based model to TD maps obtained from a single-channel ultra-wideband (UWB) radar to estimate 2D joint positions.
MDPose~\cite{tang2024mdpose} estimated joint velocities from TD maps of passive WiFi radar using a CNN--RNN and combined them with pose optimization to reconstruct skeletal motion.
MDST~\cite{zhou2024mdst} estimated 2D poses from TD maps through Swin Transformer-based feature learning, while Radar PoseLifter~\cite{zhou2024radarposelifter} applied temporal convolutions to estimated 2D joint sequences to transform them into 3D poses.

To compensate for the limited spatial information in TD maps and the observation-direction dependence of Doppler measurements, MVDoppler-Pose~\cite{mvdoppler_pose} combined TD and time--range (TR) maps acquired from two different observation directions.
This method estimates 3D human poses by fusing complementary TD/TR features and information across observation directions, while applying supervision to both joint positions and temporal motion.
This demonstrates that temporal signal features and range information from TD/TR maps can be jointly utilized for pose estimation.

\subsection{Datasets for Radar-Based Human Pose Estimation}
\label{subsec:related_radar_datasets}

Radar-based HPE datasets have primarily been constructed around different radar representations, including point clouds and radar tensors that provide spatial observations, as well as TD/TR maps that represent temporal variations in radar signals.
Among spatial-observation-based datasets, mRI~\cite{an2022health_mri} collected 12 types of rehabilitation movements, including squats, lunges, stretching, and walking, in indoor environments and provides radar point clouds, RGB-D, and inertial measurement unit (IMU) data.
mmBody~\cite{chen2022mmbody} collected 100 types of movements, including static poses, daily activities, and sports movements, in indoor environments and simulated adverse weather and occlusion conditions, and provides radar point clouds, human skeleton annotations, and mesh annotations.
MM-Fi~\cite{yang2023mm} collected 27 types of activities, including squats, lunges, jumping, and object picking, across various indoor environments and provides synchronized radar point clouds, LiDAR, RGB-D, and WiFi data.
MiliPoint~\cite{cui2023milipoint} provides point clouds and 3D joint annotations for 49 types of aerobic exercises performed indoors.
Subsequently, RT-Pose~\cite{ho2024rt} provided analog-to-digital converter (ADC) data, 4D radar tensors, and 3D joint annotations for six types of activities in indoor and outdoor environments, while M4Human~\cite{fan2026m4human} provides spatial radar tensors, point clouds, human mesh annotations, and joint annotations for 50 types of activities, including rehabilitation and sports.
DualRadarHPE~\cite{nguyen2026multi} synchronized ADC data, radar heatmaps, and point clouds with 3D joint GT for full-body movements, object manipulation, and multi-person scenarios in indoor and outdoor environments.
These datasets primarily support human pose estimation research using 3D spatial observations from radar, while some also provide raw ADC data that enable additional signal processing.

TD-based datasets that utilize temporal Doppler information include HPSUR~\cite{zhou2024radarposelifter} and MVDoppler-Pose~\cite{mvdoppler_pose}.
HPSUR collected four types of activities, including walking, punching, sit-to-stand transitions, and falling, in indoor environments and provides TD maps obtained using a single-input single-output (SISO) UWB radar and 3D joint GT acquired through inertial motion capture.
MVDoppler-Pose collected data using two frequency-modulated continuous-wave (FMCW) radars and a stereo camera under various walking trajectories and self-occlusion conditions, and provides TD/TR maps from two observation directions and camera-based 3D joint GT.
Neither dataset has been confirmed to provide radar point clouds as spatial observation inputs.

Although existing datasets provide spatial observations and temporal signal representations in various forms, publicly available datasets that jointly provide TD/TR maps, 3D radar point clouds, and 3D joint GT from the same motion recordings, enabling the evaluation of individual and fused radar inputs for HPE under identical conditions, remain limited.

\begin{figure*}[t]
    \centering
    \includegraphics[width=\textwidth]{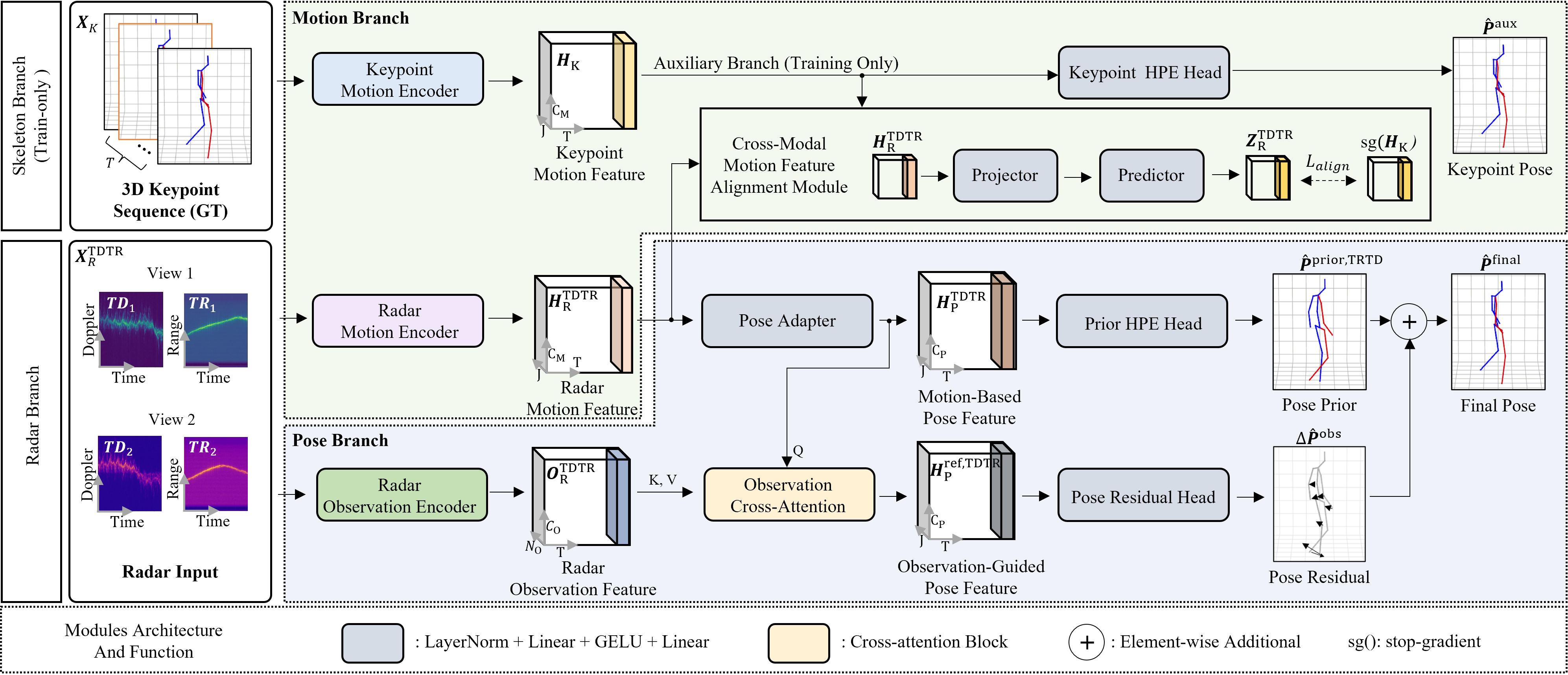}
    \caption{Overview of the I2R-Pose framework.
    The TD/TR-only configuration with two observation views is illustrated as an example.
    The Motion Branch learns joint-wise features from radar observations,
    while the Pose Branch sequentially performs initial pose estimation
    using these features and pose refinement using separately extracted observation features.
    The GT-based Keypoint Motion Encoder and Keypoint HPE Head are used only during training,
    and Cross-Modal Motion Feature Alignment guides the learning of joint-wise radar features.
    For visual clarity, the feature tensors are illustrated with the temporal dimension ($T$)
    along the X-axis, the joint or token dimension ($J$ or $N_{\mathrm O}$)
    along the Y-axis, and the channel dimension ($C$) along the Z-axis.}
    \label{fig:overall_framework}
\end{figure*}

\begin{figure}[ht]
    \centering
    \subfloat[]{\includegraphics[width=0.98\columnwidth]{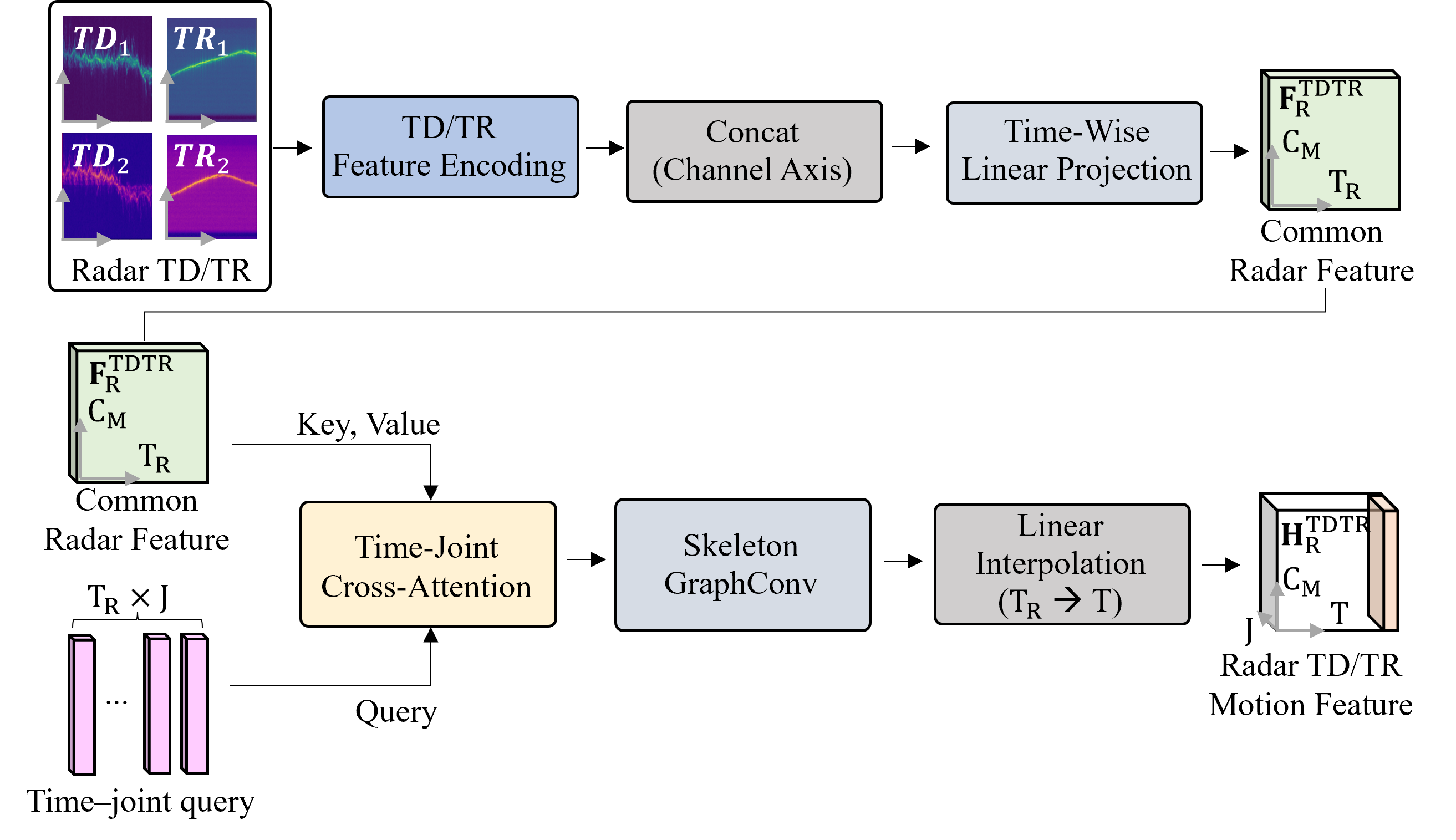}\label{fig:radar_motion_encoder}}
    \par\vspace{3pt}
    \subfloat[]{\includegraphics[width=0.98\columnwidth]{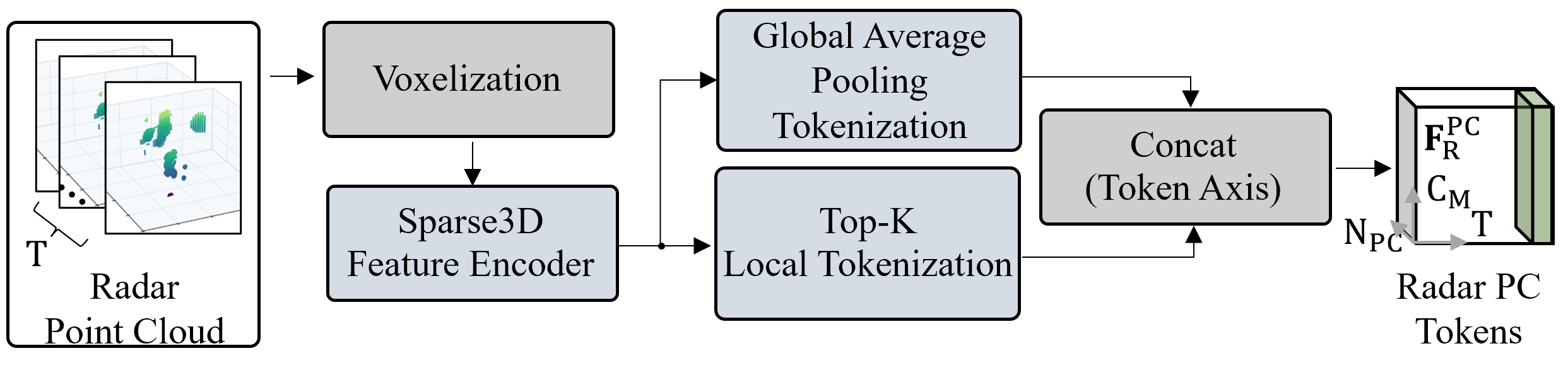}\label{fig:pc_observation_encoder}}
    \par\vspace{3pt}
    \subfloat[]{\includegraphics[width=0.98\columnwidth]{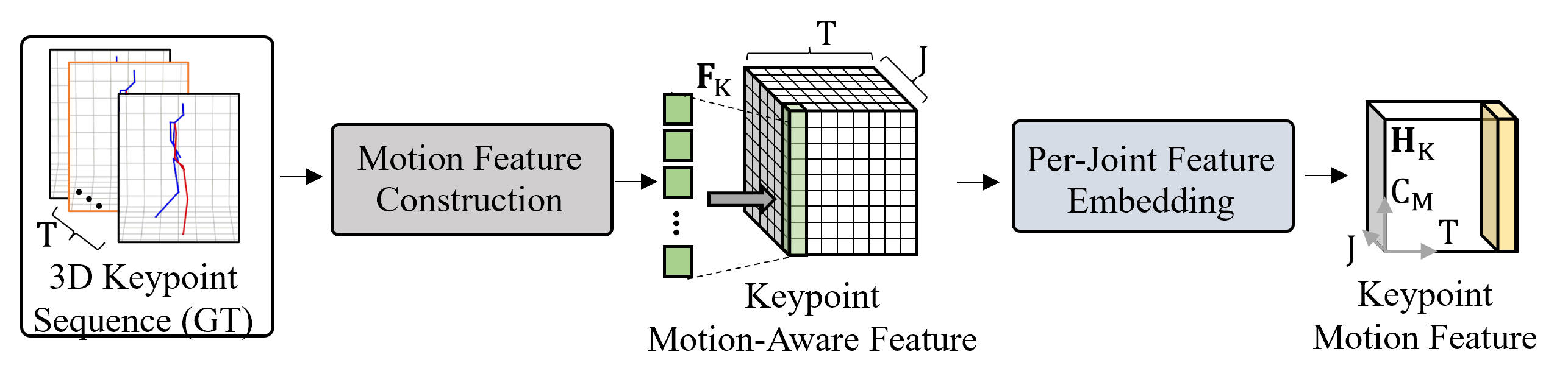}\label{fig:keypoint_motion_encoder}}
    \par\vspace{3pt}
    \caption{Main feature extraction and representation processes in I2R-Pose.
    (a) Generation of joint-wise radar motion features
    through TD/TR feature extraction and Time--Joint Cross-Attention.
    (b) Generation of PC spatial tokens through voxel-based
    feature extraction and global and local tokenization.
    The illustrated motion features are generated in the
    Radar Motion Encoder, while the same tokenization
    structure is independently used in the Radar Observation
    Encoder with separate weights.
    (c) Construction of joint-wise motion descriptors
    from GT joint sequences and their transformation
    into keypoint motion features.}
    \label{fig:encoder_architectures}
\end{figure}
\section{Method}
\label{sec:method}

In this study, we propose \textit{I2R-Pose (Initialization-to-Refinement Pose Framework)}, which estimates a motion-aware initial pose from radar observations and refines it using separately extracted observation features.
Fig.~\ref{fig:overall_framework} illustrates the overall two-stage architecture using the TD/TR-only configuration as an example.
The proposed method consists of \textit{Motion-Aware Pose Initialization} and \textit{Observation-Guided Refinement}.
The former encompasses both joint-wise feature learning in the Motion Branch and initial pose prediction in the Pose Branch, while the latter refers to refinement based on separately extracted observation features in the Pose Branch.
The same sequential estimation principle is applied to the PC-only and TD/TR--PC fusion configurations.

\subsection{Overall Framework}
\label{subsec:overall_framework}

The radar input consists of time--Doppler (TD) and time--range (TR) maps from $V$ observation views, denoted as $\mathcal X_{\mathrm R}=\{(\mathbf X_v^{\mathrm{TD}},\mathbf X_v^{\mathrm{TR}})\}_{v=1}^{V}$, and 3D radar point clouds $\mathbf X_{\mathrm{PC}}$ when available.
I2R-Pose considers TD/TR-only, PC-only, and TD/TR--PC fusion inputs, with different paths for generating joint-wise features depending on the available radar representations.
Although all input configurations share the same sequential estimation principle, they are trained separately using input-specific encoders.
The final output is $\widehat{\mathbf P}^{\mathrm{final}}\in\mathbb R^{B\times T\times J\times3}$, where $B$, $T$, and $J$ denote the batch size, number of output time steps, and number of joints, respectively.
The predicted poses are represented as root-relative coordinates following the GT coordinate convention of each input dataset.

The \emph{Motion Branch} transforms radar observations into joint-wise features $\mathbf H_{\mathrm R}\in\mathbb R^{B\times T\times J\times C_{\mathrm M}}$.
During training, keypoint motion features derived from GT joint sequences are used to guide joint-wise feature learning.
The \emph{Pose Branch} predicts an initial pose $\widehat{\mathbf P}^{\mathrm{prior}}$ from these features and generates the final pose by estimating positional corrections $\Delta\widehat{\mathbf P}^{\mathrm{obs}}$ from separate observation features.
The GT-based auxiliary path is not used during inference.

\subsection{Motion Branch}
\label{subsec:Motion_Branch}

The Motion Branch constructs joint-wise features by utilizing temporal relationships in radar inputs and human skeletal structure.
Coordinate regression for the initial pose is performed in the subsequent Pose Branch.
Although TD/TR and PC inputs undergo different feature encoding processes, both are converted into the same joint-wise feature format for pose estimation.
The main feature extraction and representation processes
used in I2R-Pose are illustrated in
Fig.~\ref{fig:encoder_architectures}.

\subsubsection{Radar Motion Encoder}
\label{subsubsec:Radar_motion_encoder}

The Radar Motion Encoder transforms radar observation features into joint-wise motion features using different feature extraction paths for TD/TR-only, PC-only, and TD/TR--PC fusion input configurations.

TD/TR inputs are processed through TD/TR Feature Encoding, as illustrated in Fig.~\ref{fig:radar_motion_encoder}.
We adopt the modality- and view-attention-based fusion architecture of MVDoppler-Pose~\cite{mvdoppler_pose}, while employing independent lightweight ResNet-18-based encoders~\cite{resnet} for each TD/TR input to hierarchically extract local signal patterns along the Doppler and range axes.
During this process, the temporal resolution is preserved, and signal-axis attention pooling selectively aggregates features along the Doppler or range axis at each time step.
This produces TD features $\mathbf F_{\mathrm{TD}}\in\mathbb R^{B\times512\times128}$ and TR features $\mathbf F_{\mathrm{TR}}\in\mathbb R^{B\times128\times128}$.
The TR features are then linearly interpolated to match the TD temporal length of 512, followed by modality attention within each observation view and view attention across multiple observation views.
The features from each stream are concatenated along the channel dimension and processed through Time-Wise Linear Projection to generate common TD/TR radar features $\mathbf F_{\mathrm R}^{\mathrm{TDTR}} \in\mathbb R^{B\times T_{\mathrm R}\times C_{\mathrm M}}$, where $T_{\mathrm R}=512$ and $C_{\mathrm M}=128$.

In the TD/TR-only configuration, Time--Joint Cross-Attention transforms the common radar features into joint-wise representations.
To this end, time--joint queries are constructed by combining learnable joint-wise queries, joint identity embeddings, time embeddings, and radar context at the current time step.
The resulting queries are represented as $\mathbf Q\in\mathbb R^{B\times(T_{\mathrm R}J)\times C_{\mathrm M}}$, while the common TD/TR radar features $\mathbf F_{\mathrm R}^{\mathrm{TDTR}}$, augmented with separate time embeddings, form the key/value inputs $\mathbf M\in\mathbb R^{B\times T_{\mathrm R}\times C_{\mathrm M}}$.
Time--Joint Cross-Attention allows each joint query to attend to radar features across the entire temporal interval.
Subsequently, skeleton graph convolution is applied to incorporate inter-joint connectivity, producing $\mathbf H_{\mathrm R}^{\mathrm{base}} \in\mathbb R^{B\times T_{\mathrm R}\times J\times C_{\mathrm M}}$.
Finally, the temporal dimension is linearly interpolated to match the output pose time steps $T$, yielding $\mathbf H_{\mathrm R}^{\mathrm{TDTR}} \in\mathbb R^{B\times T\times J\times C_{\mathrm M}}$.

In the PC-only configuration, spatial tokens are generated through the voxel-based feature extraction and tokenization process illustrated in Fig.~\ref{fig:pc_observation_encoder}. Each radar point contains its 3D coordinates, Doppler value, and radar power value. After voxelization, normalized point features and voxel-relative offsets are embedded using a point MLP and aggregated within each occupied voxel. A sparse 3D convolutional encoder extracts spatial features by aggregating information from neighboring occupied voxels. Global and local tokens are then generated through the Global Average Pooling Tokenization and Top-K Local Tokenization blocks, respectively. The global token is obtained by adding a learnable bias to the mean of valid voxel features, while up to 96 local tokens are selected according to learned importance scores and augmented with global information. Unavailable local tokens are padded and masked. This produces PC spatial tokens $\mathbf F_{\mathrm R}^{\mathrm{PC}}\in\mathbb R^{B\times T\times N_{\mathrm{PC}}\times C_{\mathrm M}}$, where $N_{\mathrm{PC}}=97$ consists of one global token and up to 96 local tokens, and $C_{\mathrm M}=128$ denotes the feature dimension.

Subsequently, joint-wise queries attend to the PC spatial
tokens $\mathbf F_{\mathrm R}^{\mathrm{PC}}$
through cross-attention, following the joint-wise
attention mechanism illustrated in
Fig.~\ref{fig:radar_motion_encoder}.
This produces joint-wise features
$\mathbf H_{\mathrm{PC}}^{\mathrm{joint}}
\in\mathbb R^{B\times T\times J\times C_{\mathrm M}}$.
Skeleton graph convolution is then applied to incorporate
inter-joint connectivity, yielding the PC-only motion features
$\mathbf H_{\mathrm R}^{\mathrm{PC}}
\in\mathbb R^{B\times T\times J\times C_{\mathrm M}}$,
which are used for initial pose estimation.

In the TD/TR--PC fusion configuration, temporal motion information from TD/TR maps is additionally incorporated into the joint-wise features generated from PCs.
To this end, cross-attention is performed using $\mathbf H_{\mathrm{PC}}^{\mathrm{joint}}$ as queries and the common TD/TR radar features $\mathbf F_{\mathrm R}^{\mathrm{TDTR}}$ as keys/values.
Skeleton graph convolution is then applied to generate joint-wise features incorporating TD/TR information, $\mathbf H_{\mathrm R}^{\mathrm{cross}} \in\mathbb R^{B\times T\times J\times C_{\mathrm M}}$.
These features are added element-wise to the PC joint-wise features, followed by an additional skeleton graph convolution and normalization to obtain the final fused features $\mathbf H_{\mathrm R}^{\mathrm{Fusion}} \in\mathbb R^{B\times T\times J\times C_{\mathrm M}}$.

The final motion features from each input configuration are used for initial pose estimation in the Pose Branch.
In the fusion configuration, the final fused features $\mathbf H_{\mathrm R}^{\mathrm{Fusion}}$ are passed to the Pose Branch, while the TD/TR-referenced features $\mathbf H_{\mathrm R}^{\mathrm{cross}}$ are used for motion feature alignment during training.


\begin{table}[t]
\centering
\caption{Composition of the 52-dimensional keypoint motion descriptor.}
\label{tab:motion_features}
\IIRTableStyle

\begin{tabularx}{\columnwidth}{@{}>{\raggedright\arraybackslash}p{0.32\columnwidth}>{\raggedright\arraybackslash}X>{\centering\arraybackslash}p{0.075\columnwidth}@{}}
\toprule
Feature group & Components & Dim. \\
\midrule
Body geometry & Root-relative positions and bone vectors & 6 \\
Temporal changes & Joint, bone, and parent-relative differences & 18 \\
Motion statistics & Motion magnitudes and reference-axis differences & 4 \\
Multiscale relations & Root and bone temporal cross-products & 24 \\
\midrule
\textbf{Total} & & \textbf{52} \\
\bottomrule
\end{tabularx}
\end{table}

\subsubsection{Keypoint Motion Encoder}
\label{subsubsec:Keypoint_motion_encoder}

The training-only Keypoint Motion Encoder generates a 52-dimensional joint-wise descriptor $\mathbf F_{\mathrm K}\in\mathbb R^{B\times T\times J \times 52}$ from the GT joint sequence $\mathbf X_{\mathrm K}=\mathbf P^{\mathrm{gt}}$ through Motion Feature Construction, as illustrated in Fig.~\ref{fig:keypoint_motion_encoder}.
The descriptor captures both body geometry and temporal changes and consists of root-relative positions and bone information, temporal differences, motion magnitudes, and relationships across multiple temporal intervals (Table~\ref{tab:motion_features}). Further details are provided in Appendix~\ref{app:keypoint_descriptor}.
Per-Joint Feature Embedding applies a shared feed-forward network (FFN) and joint embeddings to obtain $\mathbf H_{\mathrm K}\in\mathbb R^{B\times T\times J \times C_{\mathrm M}}$.
A separate Keypoint HPE Head is trained to predict $\widehat{\mathbf P}^{\mathrm{aux}}$ from $\mathbf H_{\mathrm K}$.

\subsubsection{Cross-Modal Motion Feature Alignment}
\label{subsubsec:Cross-Modal_Motion_Latent_Alignment}

To facilitate the learning of radar-based joint-wise features, a projector $g_{\mathrm R}$ and a predictor $q_{\mathrm R}$ are applied to the joint-wise features generated from each input configuration, and the resulting features are aligned with the GT-based features $\mathbf H_{\mathrm K}$ at corresponding time steps and joints.
\begin{equation}
\begin{aligned}
\mathbf Z_{\mathrm R}
&=q_{\mathrm R}\!\left(g_{\mathrm R}(\widetilde{\mathbf H}_{\mathrm R})\right),\\
\mathbf Z_{\mathrm K}
&=\operatorname{sg}(\mathbf H_{\mathrm K}).
\end{aligned}
\label{eq:teacher_student_representation}
\end{equation}
Here, $\widetilde{\mathbf H}_{\mathrm R}$ denotes $\mathbf H_{\mathrm R}^{\mathrm{TDTR}}$ for the TD/TR-only configuration, $\mathbf H_{\mathrm R}^{\mathrm{PC}}$ for the PC-only configuration, and $\mathbf H_{\mathrm R}^{\mathrm{cross}}$ for the TD/TR--PC fusion configuration. These features are aligned to the $T$ output pose time steps, and $\operatorname{sg}$ denotes the stop-gradient operation applied to the GT-based features. The Keypoint Motion Encoder is trained using an auxiliary pose loss, while the final radar-based joint-wise features $\mathbf H_{\mathrm R}$ are passed to the Pose Branch without passing through the projector or predictor. The radar-side predictor maps the projected radar features toward the GT-based motion features.

\subsection{Pose Branch}
\label{subsec:Pose_Branch}

The Pose Branch predicts an initial pose from the joint-wise features generated by the Motion Branch and estimates positional corrections using separately extracted radar observation features.

\subsubsection{Motion-Aware Pose Initialization}
\label{subsubsec:motion_based_pose_prior}

The Pose Adapter transforms the joint-wise features $\mathbf H_{\mathrm R}$ into $\mathbf H_{\mathrm P}\in\mathbb R^{B\times T\times J \times C_{\mathrm P}}$.
The Prior HPE Head predicts the initial pose $\widehat{\mathbf P}^{\mathrm{prior}}$ from these features.
$\mathbf H_{\mathrm P}$ is also used as the query for subsequent observation-guided pose refinement.

\subsubsection{Radar Observation Encoder}
\label{subsubsec:Radar_Observation_encoder}

The Radar Observation Encoder independently generates observation tokens $\mathbf O_{\mathrm R}\in\mathbb R^{B\times T\times N_{\mathrm O}\times C_{\mathrm O}}$ using input-specific feature extraction processes.
The TD/TR and PC observation paths employ the corresponding feature extraction and tokenization structures used in the Motion Branch, but do not share encoder weights with it.

In the TD/TR-only configuration, observation tokens are generated using the TD/TR feature extraction structure described in Section~\ref{subsubsec:Radar_motion_encoder}.
For the two-view setting, four stream-specific tokens and one fused token are constructed at each time step, yielding $N_{\mathrm O}=5$.

In the PC-only and TD/TR--PC fusion configurations, the PC feature extraction and tokenization process illustrated in Fig.~\ref{fig:pc_observation_encoder} is applied independently to generate observation tokens $\mathbf O_{\mathrm R}^{\mathrm{PC}}\in\mathbb R^{B\times T\times N_{\mathrm{PC}}\times C_{\mathrm O}}$.
Here, $N_{\mathrm{PC}}=97$ consists of one global token and up to 96 local tokens, and $C_{\mathrm O}=128$ denotes the observation feature dimension.

In all input configurations, the resulting observation tokens are used as keys and values for observation-guided pose refinement.

\subsubsection{Observation Cross-Attention}
\label{subsec:observation_guided_refinement}

At each time step, cross-attention is performed using $\mathbf H_{\mathrm P}$ as queries and $\mathbf O_{\mathrm R}$ as keys/values to generate joint-wise features $\mathbf H_{\mathrm P}^{\mathrm{ref}}$ incorporating observation information.
The Pose Residual Head predicts 3D positional corrections for each joint from these features and adds them to the initial pose.
\begin{equation}
\begin{aligned}
\Delta\widehat{\mathbf P}^{\mathrm{obs}}
&=\alpha\tanh\!\left(f_{\mathrm{res}}(\mathbf H_{\mathrm P}^{\mathrm{ref}})\right),\\
\widehat{\mathbf P}^{\mathrm{final}}
&=\widehat{\mathbf P}^{\mathrm{prior}}+\Delta\widehat{\mathbf P}^{\mathrm{obs}}.
\end{aligned}
\label{eq:pose_refinement}
\end{equation}
Here, $\alpha$ is a constant that controls the correction range for each coordinate component, and $\alpha=0.1\,\mathrm m$ is used in this study.
This refinement does not directly compare the initial pose with radar signals based on physical principles; rather, it predicts residual corrections by referring to separately extracted observation features.

\begin{table*}[t]
\centering
\caption{Comparison of radar-based HPE datasets providing 3D joint GT.}
\label{tab:radar_hpe_datasets}
\IIRTableStyle
\setlength{\tabcolsep}{2.5pt}

\begin{tabularx}{\textwidth}{@{}
>{\raggedright\arraybackslash}p{2.12cm}
>{\centering\arraybackslash}p{0.75cm}
>{\raggedright\arraybackslash}p{1.55cm}
>{\raggedright\arraybackslash}X
>{\raggedright\arraybackslash}p{2.55cm}
>{\raggedright\arraybackslash}p{2.80cm}
>{\raggedright\arraybackslash}p{3.30cm}
@{}}
\toprule
Dataset & Year & Data &
Pose and activity coverage & Size &
Radar & Main characteristics \\
\midrule

MARS~\cite{an2021mars} &
2021 & PC &
Arm and leg exercises, squats, lunges, and other rehabilitation activities (10 types) &
40,083 frames, approx. 70 min &
TI IWR1443BOOST &
Rehabilitation-focused; PC representation \\
\addlinespace[1.5pt]

mRI~\cite{an2022health_mri} &
2022 & PC &
Squats, lunges, stretching, walking, and other rehabilitation activities (12 types) &
160k+ frames &
TI IWR1443BOOST &
Rehabilitation-focused; PC representation \\
\addlinespace[1.5pt]

mmBody~\cite{chen2022mmbody} &
2022 & PC &
Static poses, daily activities, exercise, and sports &
200k+ frames &
Arbe Phoenix &
PC-based 3D human body reconstruction \\
\addlinespace[1.5pt]

HIBER~\cite{wu2022rfmask} &
2022 & HM &
Standing, sitting, walking, and squatting &
804,760 frames &
TI AWR2243 Cascade $\times2$ &
radio-frequency heatmap-based pose and activity observations \\
\addlinespace[1.5pt]

MM-Fi~\cite{yang2023mm} &
2023 & PC &
Squats, lunges, jumping, object picking, and other daily and rehabilitation activities (27 types) &
320,760 frames, 1,080 sequences &
TI IWR6843AOP &
Diverse activities and PC-based HPE \\
\addlinespace[1.5pt]

MiliPoint~\cite{cui2023milipoint} &
2023 & PC &
Arm- and leg-focused aerobic exercises (49 types) &
545,059 frames &
TI IWR1843 &
PC-based evaluation of diverse movements \\
\addlinespace[1.5pt]

HPSUR~\cite{zhou2024radarposelifter} &
2024 & TD &
Walking, punching, sit-to-stand transitions, and falling (4 types) &
311,963 frames, 5 subjects &
SISO UWB radar\newline (2.7--3.2 GHz) &
Single-view TD with 17-joint GT \\
\addlinespace[1.5pt]

RT-Pose~\cite{ho2024rt} &
2024 & ADC, 4D RT &
Walking, sitting, arm and leg exercises, and other activities (6 types) &
72,000 frames, 240 sequences &
TI AWR2243 Cascade &
ADC- and radar tensor-based HPE \\
\addlinespace[1.5pt]

mmRadPose\newline\cite{engel2025advanced} &
2025 & Raw cube, PC &
Arm and leg exercises and other activities (12 types) &
203,149 frames, 432 sequences, approx. 3.77 h &
TI IWR6843AOP-EVM &
Radar cube and PC representations \\
\addlinespace[1.5pt]

MVDoppler-\newline Pose~\cite{mvdoppler_pose} &
2025 & TD, TR &
Seven walking trajectories with changes in direction and arm movements &
6.3 h &
TI AWR1843 $\times2$ &
Multi-view TD/TR-based walking HPE \\
\addlinespace[1.5pt]

M4Human~\cite{fan2026m4human} &
2026 & RT, PC &
Daily, rehabilitation, and sports activities (50 types) &
approx. 661k frames, 999 sequences, approx. 15.3 h &
High-resolution mmWave radar\newline (model not identified) &
RT/PC with motion-capture-based 3D mesh and joint annotations \\
\addlinespace[1.5pt]

DualRadar-\newline HPE~\cite{nguyen2026multi} &
2026 & ADC, HM, PC &
Full-body movement, locomotion, and object manipulation &
52,475 frames, 40 sequences &
TI AWR2243 Cascade,\newline Bosch Off-Highway Premium &
Multiple representations: raw ADC, heatmaps, and PCs \\
\midrule

\textbf{RaMoS-}\newline\textbf{Pose (Ours)} &
\textbf{2026} &
\textbf{ADC, TD/TR, PC} &
\textbf{Static poses (standing, sitting, lying), pose transitions, locomotion (walking, running), exercise, daily activities, and simulated falls} &
\textbf{187,156 valid GT frames, 1,751 sequences, approx. 5.20 h} &
\textbf{TI MMWCAS-RF/DSP-EVM,\newline SRS RETINA-4SN} &
\textbf{Synchronized TD/TR, radar PCs, and 3D GT for diverse indoor activities} \\

\bottomrule
\end{tabularx}

\vspace{2pt}
\begin{minipage}{\textwidth}
\footnotesize
PC: radar point cloud; HM: heatmap; RT: radar tensor;
N/A: not reported or not identified.
\end{minipage}
\end{table*}

\subsection{Training Objectives}
\label{subsec:training_objectives}

I2R-Pose employs pose supervision for the initial pose, final pose, and training-only auxiliary pose predicted by the GT-based Keypoint Motion Encoder, together with motion feature alignment between radar and keypoint representations.

The pose supervision loss consists of the mean per-joint position error (MPJPE), which measures the average Euclidean distance between predicted and GT joint positions, and the temporal motion correlation loss~\cite{mvdoppler_pose}, which evaluates the consistency of temporal joint movements.
For temporal intervals $\tau\in\{2,4,6,8\}$, the motion correlation loss constructs a motion descriptor $\mathbf m_{\tau,t,j}=\mathbf p_{t,j}\times\mathbf p_{t+\tau,j}$ by taking the cross-product of 3D joint positions at two different time steps.
The temporal Pearson correlation coefficients between the descriptors obtained from the predicted and GT poses are then compared.
This allows the model to learn not only joint position accuracy but also the temporal movement patterns of arm and leg joints~\cite{mvdoppler_pose}.

The pose supervision loss is defined as
\begin{equation}
\begin{aligned}
\mathcal L_{\mathrm{pose}}(\widehat{\mathbf P})={}&
\mathcal L_{\mathrm{MPJPE}}(\widehat{\mathbf P})\\
&+0.05\bigl(
\mathcal L_{\mathrm{motion}}^{\mathrm{arm}}(\widehat{\mathbf P})
+\mathcal L_{\mathrm{motion}}^{\mathrm{leg}}(\widehat{\mathbf P})
\bigr).
\end{aligned}
\label{eq:pose_loss}
\end{equation}
Here, $\mathcal L_{\mathrm{motion}}^{\mathrm{arm}}$ and
$\mathcal L_{\mathrm{motion}}^{\mathrm{leg}}$ denote the correlation losses computed from the motion descriptors of arm and leg joints, respectively.
The same pose supervision loss is applied to the initial pose, final pose, and auxiliary pose predicted by the GT-based Keypoint Motion Encoder.
The detailed computation of each loss is provided in Appendix~\ref{app:training_loss}.

\subsubsection{Motion Feature Alignment and Total Objective}
\label{subsubsec:alignment_objective}

Cross-Modal Motion Feature Alignment aligns the radar-side alignment features $\mathbf Z_{\mathrm R}$ and GT-derived alignment features $\mathbf Z_{\mathrm K}$ in terms of their representations and temporal changes at corresponding time steps and joints.
To this end, the mean squared error (MSE) between the features is combined with a temporal difference loss that compares the directions and magnitudes of feature changes.
\begin{equation}
\mathcal L_{\mathrm{align}}
=\operatorname{MSE}(\mathbf Z_{\mathrm R},\mathbf Z_{\mathrm K})
+\lambda_\Delta\mathcal L_\Delta.
\label{eq:canonical_alignment_loss}
\end{equation}
Here, $\mathcal L_\Delta$ consists of a cosine loss that compares the directions of feature changes between adjacent time steps and a SmoothL1 loss that compares their magnitudes.
The parameter $\lambda_\Delta$ controls the weight of the temporal feature alignment loss.
The detailed computation and training schedule are described in Appendix~\ref{app:training_loss}.

The total training objective is defined as
\begin{equation}
\begin{aligned}
\mathcal L_{\mathrm{total}}={}&
\mathcal L_{\mathrm{pose}}(\widehat{\mathbf P}^{\mathrm{final}})
+\mathcal L_{\mathrm{pose}}(\widehat{\mathbf P}^{\mathrm{prior}})\\
&+\mathcal L_{\mathrm{pose}}(\widehat{\mathbf P}^{\mathrm{aux}})
+\mathcal L_{\mathrm{align}}.
\end{aligned}
\label{eq:total_loss}
\end{equation}
During inference, the GT-based Keypoint Motion Encoder, Keypoint HPE Head, and feature alignment path are removed, and the final pose is estimated using only radar inputs.

\section{RaMoS-Pose Dataset}
\label{sec:ramos_dataset}

\subsection{Motivation and Dataset Comparison}
\label{subsec:ramos_motivation}

To apply radar-based 3D human pose estimation (HPE) to indoor living environments, it is necessary to evaluate performance across diverse motions, including not only walking but also static poses, pose transitions, and daily activities. In addition, time--Doppler (TD) and time--range (TR) maps provide temporal motion information and range distributions, respectively, while radar point clouds (PCs) provide the 3D positions of observed reflection points. Therefore, it is necessary to compare the individual performance of these inputs and the complementary effects of their fusion under the same motions.

Table~\ref{tab:radar_hpe_datasets} compares major radar-based HPE datasets providing 3D joint GT. Existing datasets offer different advantages in terms of motion coverage and radar representations, and some provide raw ADC signals or radar tensors together with PCs. In this study, we construct \textit{Radar Motion and Spatial Information for 3D Human Pose Estimation (RaMoS-Pose)} by synchronizing TD/TR maps and PCs acquired from a separate radar to enable comparisons under the same conditions. RaMoS-Pose includes static poses, pose transitions, locomotion, and daily activities, and provides synchronized TD/TR maps, PCs, and 17-joint GT.

\subsection{Sensor Configuration and Ground Truth}
\label{subsec:ramos_sensors}

RaMoS-Pose consists of time--Doppler/time--range maps, 3D radar point clouds, and synchronized 3D GT for 17 joints. Complex ADC signals for generating TD/TR maps are collected using the TI MMWCAS-RF/DSP-EVM equipped with four AWR2243 chips, while point clouds are acquired from a separate SRS RETINA-4SN imaging radar. By collecting different radar representations of the same motions, we analyze the complementarity between temporal signal features and spatial reflection-point information. Radar point clouds acquired from the SRS RETINA-4SN are transformed into the TI radar coordinate system using calibrated sensor extrinsic parameters.

\begin{figure}[t]
    \centering
    \includegraphics[width=\columnwidth]{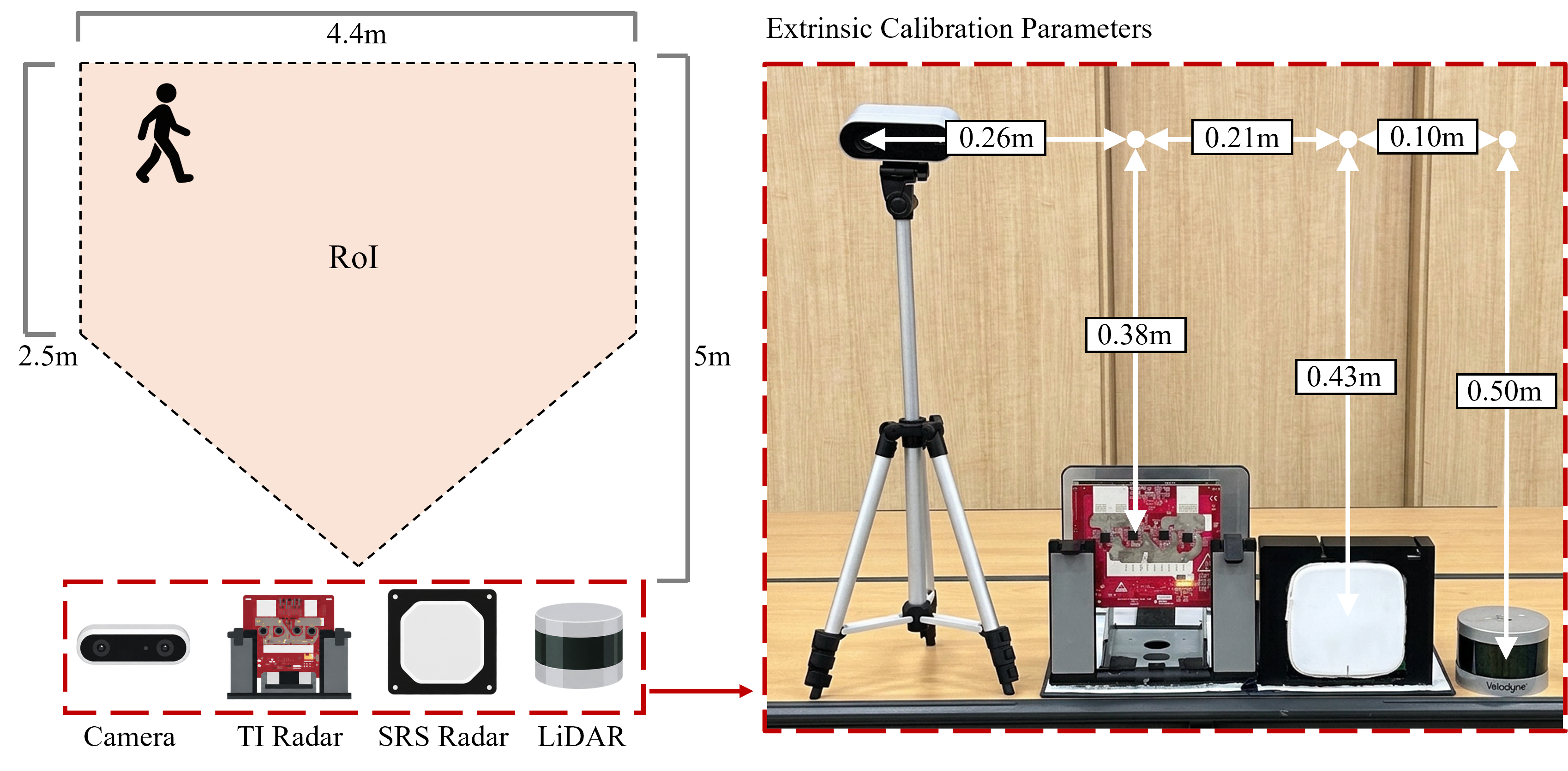}
    \caption{Data collection area and sensor configuration of RaMoS-Pose. The left panel shows the radar, camera, and LiDAR arrangement and the data collection area, while the right panel shows the actual sensor setup and relative distances between sensors.}
    \label{fig:ramos_sensor_setup}
\end{figure}

The GT is constructed based on initial 3D joint positions obtained using the skeleton-tracking SDK provided for the ORBBEC Femto Bolt RGB-D camera. Joint positions requiring correction are manually refined with reference to RGB-D images and LiDAR observations, and the joint positions and radar observations are aligned to a common coordinate system using sensor extrinsic parameters. Pelvis-relative joint coordinates are used for model training and evaluation. The camera and LiDAR are used for GT generation and alignment but are not used as inference inputs to I2R-Pose. Fig.~\ref{fig:ramos_sensor_setup} shows the data collection area and sensor configuration.

\begin{table}[ht]
\centering
\caption{Distribution of representative motion categories in RaMoS-Pose. The 1,751 sequences are classified into mutually exclusive categories based on their primary motion, and percentages are calculated based on the number of sequences.}
\label{tab:ramos_motion_coverage}
\IIRTableStyle

\begin{tabularx}{\columnwidth}{@{}
    >{\raggedright\arraybackslash}p{1.45cm}
    >{\raggedright\arraybackslash}X >{\centering\arraybackslash}p{0.99cm} >{\centering\arraybackslash}p{1.16cm}
@{}}
\toprule
Category & Included activities & Sequences & Ratio (\%) \\
\midrule
Static poses & Standing, sitting, lying, leaning against a wall & 335 & 19.13 \\
Pose transitions & Sitting down, standing up, lying down, getting up from a lying position & 60 & 3.43 \\
Locomotion & Walking, running & 296 & 16.90 \\
Exercise & Exercise-labeled motions, push-ups, squats, jumping & 84 & 4.80 \\
Daily activities & Cleaning, mobile phone use and calls, object manipulation and organization, moving furniture, desk work, drinking, eating & 877 & 50.09 \\
Falls & Simulated falls & 27 & 1.54 \\
Free and complex motions & Free motions, complex motions & 72 & 4.11 \\
\midrule
\textbf{Total} & & \textbf{1,751} & \textbf{100.00} \\
\bottomrule
\end{tabularx}
\end{table}

\subsection{TD/TR Acquisition and Preprocessing}
\label{subsec:ramos_tdtr}

Raw ADC data are collected at 100~ms intervals (10~Hz) using the AWR2243 cascade radar of the TI MMWCAS-RF/DSP-EVM. The FMCW chirps have a start frequency of 77~GHz and a frequency slope of 85.655~MHz/$\mu$s, with 256 ADC samples acquired per chirp at 12~MS/s. Each frame contains 128 repetitions of a 12-TX time-division multiplexed multiple-input multiple-output (TDM-MIMO) chirp cycle, spanning approximately 56.8~ms. A separate SRS RETINA-4SN radar acquires 3D point clouds at approximately 10~Hz.

For TD/TR map construction, three of the 12 transmit antennas (indices 9--11) and all 16 receive antennas are selected, yielding 48 TX--RX channels. A 256-point range fast Fourier transform (FFT) is applied to the selected ADC signals, and the sequence-wise mean is removed to suppress static background reflections. The resulting signals are arranged on a timestamp-based slow-time grid at intervals of approximately 0.444~ms, with unobserved samples set to zero and masked. TD maps are generated using a 64-sample Hamming window and a 128-point Doppler FFT with zero-padding, with spectral power averaged over a fixed range ROI of 0.5--6.0~m. TR maps are generated by aggregating reflected power within a 128-sample Hamming window using the first 128 range FFT bins (indices 0--127), covering approximately 0--10.43~m. For a 10-s segment, TD and TR maps are generated at 1,985 and 250 temporal positions, respectively. The window centers are uniformly spaced in time and mapped to the nearest slow-time grid positions, with TD windows advancing by 11 or 12 samples.

Power is normalized by the energy of the observation-masked Hamming window. Windows containing fewer than 15 observed samples for TD or 30 for TR, or having insufficient valid window energy, are marked invalid and stored at $-120$~dB. Before incoherent averaging across the 48 channels, channel-wise power is normalized using the median power of within-frame reference spectrograms. The resulting maps are converted to logarithmic power (dB).

For model input, a common 3-s interval is selected based on timestamps, and the TD and TR maps are linearly interpolated along the temporal axis to $512 \times 128$ and $128 \times 128$ (time $\times$ Doppler/range), respectively. The radar maps and 3D joint GT are temporally aligned using acquisition timestamps. The observed-chirp fraction is provided as an additional TD input channel, and invalid TD values are masked to zero during input normalization and after temporal interpolation. Background subtraction and channel-scale estimation use sequence-level observations and are therefore performed offline. Temporal interpolation does not reconstruct missing ADC signals.

\subsection{Participants and Motion Coverage}
\label{subsec:ramos_activities}

A total of 1,751 sequences of varying durations, each lasting at least 10 seconds, were collected from seven subjects across 11 sessions. The dataset contains 187,156 frames with valid 17-joint GT, corresponding to approximately 5.20~h of observation time when calculated at 10~Hz. Table~\ref{tab:ramos_motion_coverage} presents the distribution obtained by assigning one representative motion category to each sequence. The dataset includes 877 sequences of daily activities (50.09\%), 335 sequences of static poses (19.13\%), and 296 sequences of locomotion (16.90\%), as well as pose transitions, exercise, and free and complex motions. Falls are collected as simulated motions rather than actual accidents, and some sequences include partial occlusion caused by furniture and household objects.


\section{Experiments}
\label{sec:experiments}

\begin{table*}[t]
\centering
\caption{Quantitative comparison on the MVDoppler-Pose dataset.}
\label{tab:overall_results}
\IIRTableStyle

\begin{tabular}{@{}lccccc@{}}
\toprule
\multicolumn{3}{c}{Method} & \multicolumn{3}{c}{Evaluation metrics} \\
\midrule
Model & Year & Input & MPJPE (mm) $\downarrow$ & $\rho\uparrow$ & PCK$_{150}$ (\%) $\uparrow$ \\
\midrule
MD-Pose~\cite{zhou2023mdpose} & 2023 & TD & 113.780 & 0.173 & 71.780 \\
MDPose~\cite{tang2024mdpose} & 2024 & TD & 122.240 & 0.482 & 73.670 \\
MVDoppler-Pose~\cite{mvdoppler_pose} & 2025 & TD/TR (1 view) & 87.604 & 0.499 & 82.920 \\
MVDoppler-Pose~\cite{mvdoppler_pose} & 2025 & TD/TR (2 views) & 60.650 & 0.596 & 92.680 \\
\midrule
I2R-Pose (Ours) & 2026 & TD/TR (1 view) & 72.942 & 0.514 & 87.164 \\
\textbf{I2R-Pose (Ours)} & \textbf{2026} & \textbf{TD/TR (2 views)} & \textbf{36.692} & \textbf{0.748} & \textbf{97.328} \\
\bottomrule
\end{tabular}
\end{table*}

\begin{table*}[t]
\centering
\caption{Quantitative comparison on the RaMoS-Pose dataset. TD/TR uses a single radar view, and PC denotes SRS radar point clouds.}
\label{tab:new_dataset_main}
\IIRTableStyle

\begin{tabular}{@{}lccccc@{}}
\toprule
\multicolumn{3}{c}{Method} & \multicolumn{3}{c}{Evaluation metrics} \\
\midrule
Model & Year & Input & MPJPE (mm) $\downarrow$ & $\rho\uparrow$ & PCK$_{150}$ (\%) $\uparrow$ \\
\midrule
MD-Pose~\cite{zhou2023mdpose} & 2023 & TD & 201.793 & 0.044 & 52.358 \\
MDPose~\cite{tang2024mdpose} & 2024 & TD & 227.028 & 0.091 & 42.774 \\
MVDoppler-Pose~\cite{mvdoppler_pose} & 2025 & TD/TR & 170.418 & 0.109 & 57.126 \\
VCF-Pose~\cite{vcf-pose} & 2026 & PC & 151.098 & 0.143 & 64.290 \\
MS-STPoseNet~\cite{multiscale_hpe} & 2026 & PC & 157.680 & 0.187 & 61.100 \\
\midrule
I2R-Pose (Ours) & 2026 & TD/TR & 160.664 & 0.092 & 63.092 \\
I2R-Pose (Ours) & 2026 & PC & 129.311 & 0.189 & 70.877 \\
\textbf{I2R-Pose (Ours)} & \textbf{2026} & \textbf{TD/TR--PC} & \textbf{128.237} & \textbf{0.194} & \textbf{71.934} \\
\bottomrule
\end{tabular}
\end{table*}

\subsection{Experimental Setup}
\label{subsec:experimental_setup}

\subsubsection{Datasets and Evaluation Protocol}
I2R-Pose is evaluated on the public MVDoppler-Pose dataset and our RaMoS-Pose dataset. MVDoppler-Pose~\cite{mvdoppler_pose} was collected using FMCW radars from two observation views and stereo cameras over an area of approximately $10\,\mathrm{m}\times10\,\mathrm{m}$. It contains approximately 6.3 hours of synchronized data from 13 subjects performing seven walking trajectories. RaMoS-Pose includes static poses, pose transitions, locomotion, exercise, and daily activities. Its sensor configuration and data composition are described in Section~\ref{sec:ramos_dataset}.

For MVDoppler-Pose, we follow the subject-independent evaluation protocol of the original study, using data from 10 of the 13 subjects for training and the remaining three subjects for evaluation. For RaMoS-Pose, each sequence lasts at least 10 seconds, and the dataset is divided into training and evaluation sets at the sequence level with an approximate ratio of 8:2. Although the same activity categories may appear in both sets, the training and evaluation sets contain distinct sequences with no overlap. On MVDoppler-Pose, we compare I2R-Pose with existing TD/TR-based methods and evaluate the effects of its major components. On RaMoS-Pose, we compare the TD/TR-only, PC-only, and TD/TR--PC fusion configurations on the same evaluation sequences to analyze the effects of individual radar representations and their fusion. For a fair comparison, all existing methods are independently reimplemented based on their respective papers and evaluated using the same training and evaluation splits for each dataset.

\subsubsection{Input and Model Configuration}
I2R-Pose takes approximately 3 seconds of radar observations as input and estimates 3D poses for 17 joints at 16 time steps. The input sizes of the TD and TR maps are $512\times128$ and $128\times128$, respectively. For MVDoppler-Pose, TD/TR maps from two observation views are used. For RaMoS-Pose, TD/TR maps from a single observation view and optional 3D radar PCs are used. An additional observation validity channel is incorporated into the TD input of RaMoS-Pose to indicate unobserved intervals between radar frames.

For PC inputs, the five point attributes and three voxel-relative offsets form an 8-dimensional input to the point MLP. Points within a fixed ROI of $[0,8)\times[-4,4)\times[-2,2)$~m are voxelized at a resolution of $0.1$~m along each axis. The point MLP uses channel dimensions of 48 and 64. The sparse 3D convolutional encoder consists of two stages with 64 and 96 channels, each containing two residual blocks, with $(2,2,1)$ downsampling between stages.

\subsubsection{Training Configuration}
All experiments are conducted using two NVIDIA GeForce RTX 3090 GPUs with a total batch size of 4. The initial learning rate and weight decay of the AdamW optimizer are set to $10^{-4}$ and $10^{-2}$, respectively. The learning rate is adjusted using cosine annealing with warm restarts, with an initial period of 25 epochs and a minimum learning rate of $10^{-6}$. The gradient norm is clipped to 5, and the models are trained for up to 250 epochs.

\subsection{Evaluation Metrics}
Following the evaluation metrics of MVDoppler-Pose~\cite{mvdoppler_pose}, we use mean per-joint position error (MPJPE), Pearson correlation coefficient ($\rho$), and PCK$_{150}$. MPJPE is defined as the average 3D Euclidean distance between predicted and GT joint positions.
\begin{equation}
\mathrm{MPJPE}=\frac{1}{NTJ}\sum_{n=1}^{N}\sum_{t=1}^{T}\sum_{j=1}^{J}\left\|\widehat{\mathbf p}_{n,t,j}-\mathbf p^{\mathrm{gt}}_{n,t,j}\right\|_2.
\label{eq:eval_mpjpe}
\end{equation}
Here, $N$ is the number of evaluation segments, $T=16$ is the number of output time steps, and $J=17$ is the number of joints. Both datasets are evaluated using pelvis-relative coordinates, and MPJPE is reported in millimeters. The percentage of correct keypoints (PCK) at a 150-mm threshold (PCK$_{150}$) represents the percentage of joints with position errors of 150~mm or less. The Pearson correlation coefficient $\rho$ is calculated by averaging the temporal correlations between predicted and GT joint coordinate sequences across coordinate axes, excluding the pelvis joint, which remains fixed in root-relative coordinates. MPJPE evaluates average joint position accuracy, PCK$_{150}$ measures the percentage of joints within a predefined error threshold, and $\rho$ evaluates the consistency of temporal changes in joint coordinates.

\subsection{Quantitative Results}
\label{subsec:quantitative_results}

\subsubsection{MVDoppler-Pose}
Table~\ref{tab:overall_results} presents the quantitative results on MVDoppler-Pose. Using TD/TR maps from two observation views, I2R-Pose reduces MPJPE by 39.50\% compared with MVDoppler-Pose, while improving $\rho$ and PCK$_{150}$ by 25.50\% and 5.02\%, respectively. With a single observation view, MPJPE is also reduced by 16.74\%, demonstrating the effectiveness of the proposed method using only TD/TR observations.

\begin{figure*}[t]
\centering
\includegraphics[width=\textwidth]{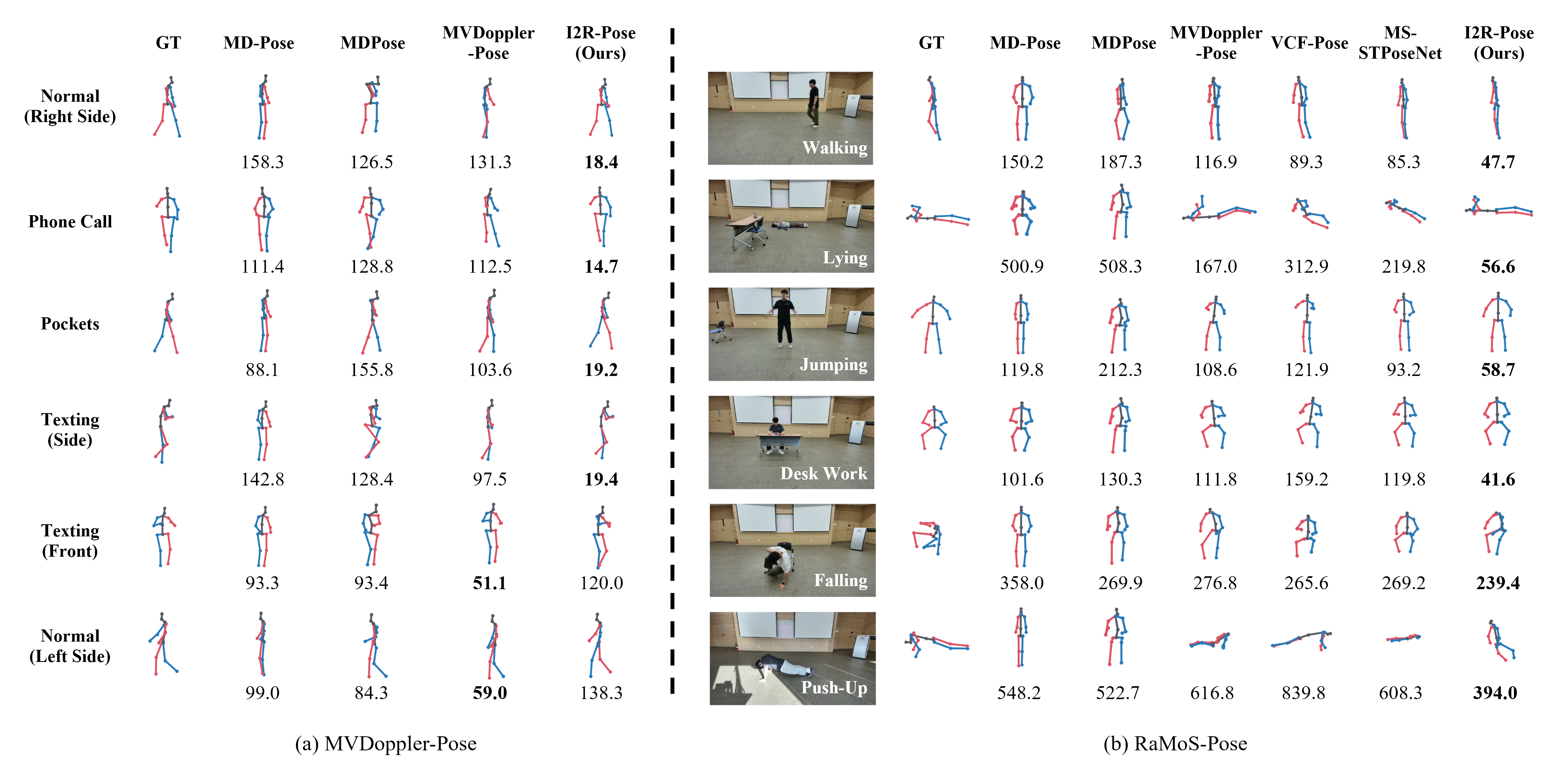}
\caption{Qualitative comparison of 3D human pose estimation results on (a) MVDoppler-Pose and (b) RaMoS-Pose. Each row shows the ground-truth (GT) pose and the poses estimated by existing methods and I2R-Pose (Ours) for a representative motion. The numbers below each estimated pose indicate the MPJPE (mm) for the corresponding frame, with bold values indicating the lowest MPJPE in each row. RGB images in (b) are provided solely for visual reference.}
\label{fig:qualitative_results}
\end{figure*}

\subsubsection{RaMoS-Pose}
On RaMoS-Pose, we compare I2R-Pose with existing TD/TR- and PC-based methods across different input configurations. As shown in Table~\ref{tab:new_dataset_main}, the TD/TR-only configuration of I2R-Pose reduces MPJPE by 5.72\% compared with the reimplemented MVDoppler-Pose, while the PC-only configuration reduces MPJPE by 14.42\% compared with VCF-Pose. The TD/TR--PC fusion configuration achieves the lowest MPJPE, with a 15.13\% reduction compared with VCF-Pose.

In particular, the improvement of the TD/TR-only configuration is smaller than the 39.50\% reduction observed on the public MVDoppler-Pose dataset. This may be attributed to the fact that RaMoS-Pose includes static poses and diverse daily activities and uses TD/TR maps from a single observation view, unlike the walking-focused public dataset. In contrast, the PC-based configurations show relatively larger performance improvements, indicating the importance of spatial information for joint position estimation across diverse indoor motions. However, the additional MPJPE reduction achieved by fusion over the PC-only configuration is limited to 0.83\%.

For further analysis, we compare the fusion configuration with the better-performing of the two existing PC-based methods (VCF-Pose and MS-STPoseNet) in each activity category. The fusion configuration reduces MPJPE by 16.78\%, 14.89\%, 16.09\%, 17.24\%, 4.39\%, and 16.94\% for static poses, locomotion, exercise, daily activities, falls, and free and complex motions, respectively. However, for pose transitions, MPJPE increases by 14.53\% compared with MS-STPoseNet. This may be attributed to rapid changes in joint configurations during pose transitions, which can make it difficult to capture temporal relationships in Motion-Aware Pose Initialization. 

The PC-only configuration contains 1.629 million inference parameters and requires $62.36 \pm 0.60$~ms per 3-s input window, whereas the TD/TR--PC fusion configuration contains 4.808 million inference parameters and requires $70.53 \pm 0.83$~ms. Both configurations were benchmarked on an NVIDIA RTX 3090 with batch size 1 and FP32, with TF32 disabled. Latency is reported as the mean and standard deviation over 100 synchronized forward passes on the same recorded input window after 30 warm-up iterations, excluding data loading, radar preprocessing, and host-to-device transfers. 

\subsection{Qualitative Results}
\label{subsec:qualitative_results}

\subsubsection{MVDoppler-Pose}
Fig.~\ref{fig:qualitative_results}(a) compares qualitative pose estimation results on MVDoppler-Pose. In the upper four examples, I2R-Pose estimates the relative arrangement of arm and leg joints during walking more closely to the GT. In contrast, the lower two examples show errors in which the model fails to sufficiently reflect changes in arm positions or estimates the forward--backward positions of the feet at an incorrect walking phase. These results suggest that, although temporal motion information from TD/TR maps can capture overall motion patterns, it may be insufficient to distinguish precise joint positions and walking phases at specific time steps.

\begin{table*}[ht]
\centering
\caption{Ablation results of major components on MVDoppler-Pose.
RME: Radar Motion Encoder; KME: Keypoint Motion Encoder and motion feature alignment; OG: Observation Cross-Attention and prior supervision.}
\label{tab:ablation_components}
\IIRTableStyle
\begin{tabular}{@{}lccccc@{}}
\toprule
\multicolumn{2}{c}{Configuration} & \multicolumn{3}{c}{Evaluation metrics} \\ \midrule
Applied modules & KME input & MPJPE (mm) $\downarrow$ & $\rho\uparrow$ & PCK$_{150}$ (\%) $\uparrow$ \\ \midrule
Baseline & -- & 60.650 & 0.596 & 92.680 \\
+RME & -- & 39.465 & 0.734 & 97.195 \\
+KME & 52D & 56.804 & 0.660 & 93.408 \\
+OG & -- & 48.901 & 0.699 & 95.348 \\
+RME+OG & -- & 37.562 & 0.744 & 97.282 \\
+RME+OG+KME & XYZ & 37.179 & 0.745 & 97.247 \\ \midrule
\textbf{RME+OG+KME (Ours)} & \textbf{52D} & \textbf{36.692} & \textbf{0.748} & \textbf{97.328} \\ \bottomrule
\end{tabular}
\end{table*}

\begin{table*}[ht]
\centering
\caption{Ablation study of motion and observation input configurations on RaMoS-Pose.}
\label{tab:ablation_pc_fusion}
\IIRTableStyle
\begin{tabular}{@{}lccccc@{}}
\toprule
\multicolumn{3}{c}{Input configuration} & \multicolumn{3}{c}{Evaluation metrics} \\ \midrule
Experiment & Radar Motion Encoder & Radar Observation Encoder & MPJPE (mm) $\downarrow$ & $\rho\uparrow$ & PCK$_{150}$ (\%) $\uparrow$ \\ \midrule
F1 & TD/TR & TD/TR & 160.664 & 0.092 & 63.092 \\
F2 & PC & PC & 129.311 & 0.189 & 70.877 \\
F3 & TD/TR & PC & 150.882 & 0.163 & 65.784 \\
F4 & TD/TR & TD/TR--PC & 149.910 & 0.174 & 67.369 \\
F5 & TD/TR--PC & TD/TR--PC & 129.669 & \textbf{0.199} & 71.018 \\ \midrule
\textbf{F6 (Ours)} & \textbf{TD/TR--PC} & \textbf{PC} & \textbf{128.237} & 0.194 & \textbf{71.934} \\ \bottomrule
\end{tabular}
\end{table*}

\subsubsection{RaMoS-Pose}
Fig.~\ref{fig:qualitative_results}(b) presents qualitative comparisons for selected motions from RaMoS-Pose. Compared with existing methods, I2R-Pose estimates the leg arrangements during walking and running, arm and leg poses during jumping, and upper-body and arm positions during desk work and cleaning more closely to the GT. However, for simulated falls and push-ups, the model does not sufficiently reproduce the characteristic full-body poses of these motions, despite achieving lower joint position errors than existing methods. These results suggest the need for joint-wise feature learning that more accurately captures large pose transitions and diverse movements.

\subsection{Ablation Study}
\label{subsec:ablation}

\subsubsection{Effect of Major Components}
\label{subsubsec:ablation_major_components}
Table~\ref{tab:ablation_components} compares RME, KME, and OG against the MVDoppler-Pose-based Baseline to analyze the effects of the major components. The Baseline is a single-stage architecture that directly estimates poses from joint-wise features, whereas OG denotes the sequential estimation structure that predicts an initial pose and then refines it using independently extracted radar observation features. RME denotes the Radar Motion Encoder, and KME denotes the GT-based Keypoint Motion Encoder and motion feature alignment. Models incorporating KME also use auxiliary pose supervision, and the 52-dimensional motion descriptor and XYZ coordinates are compared as KME inputs.

Compared with the Baseline, applying RME, KME, and OG individually reduces MPJPE by 34.93\%, 6.34\%, and 19.37\%, respectively. RME provides the largest improvement, supporting the effectiveness of motion-aware joint-wise feature learning by extracting joint-wise features from temporal radar features through time--joint queries and incorporating skeletal relationships. Applying OG to the Baseline also reduces MPJPE by 19.37\%, demonstrating the effectiveness of sequential pose refinement combining initial pose estimation and Observation-Guided Refinement. Adding OG to RME further reduces MPJPE by 4.82\%.

Meanwhile, KME guides the learning of joint-wise radar features using GT-derived motion features, reducing MPJPE by 6.34\% when applied individually and by 2.32\% when added to RME+OG. Using the 52-dimensional motion descriptor, which incorporates temporal movements and body geometry, instead of XYZ coordinates further reduces MPJPE by 1.31\%. However, the improvement from KME remains relatively limited, suggesting that motion feature alignment may be challenging due to the different representation characteristics of radar observation features and GT-derived motion features. Therefore, more effective alignment of motion relationships between these two representations may be an important direction for future improvement. The final RME+OG+KME configuration achieves the lowest error, reducing MPJPE by 39.50\% compared with the Baseline.

\subsubsection{Radar Input Configurations for Initialization and Refinement}
\label{subsubsec:ablation_pc_fusion}

On RaMoS-Pose, we analyze the roles of TD/TR and PC by varying the input configurations of the Motion Branch, which performs initial pose estimation, and the Radar Observation Encoder, which supports pose refinement.

Table~\ref{tab:ablation_pc_fusion} compares performance across different input configurations of the Radar Motion Encoder and Radar Observation Encoder. Compared with F1, F3 and F4, which incorporate PC into the Observation Encoder, reduce MPJPE by 6.09\% and 6.69\%, respectively. Furthermore, comparing F3 and F2, which use the same PC Observation Encoder, incorporating PC into the Motion Encoder reduces MPJPE by 14.30\%. These results demonstrate that the spatial information provided by PC plays an important role in both initial pose estimation and observation-guided refinement on RaMoS-Pose.

In F6, joint-wise features are first constructed from PC, and initial pose estimation is performed by referring to temporal features from TD/TR using these PC-based joint-wise features. F6 reduces MPJPE by 0.83\% compared with the PC-only configuration F2, while increasing $\rho$ by 0.005 and PCK$_{150}$ by 1.057 percentage points. These results suggest that incorporating TD/TR information into PC-based joint-wise features can provide additional, although limited, improvements.

Meanwhile, comparing F5 and F6, which use the same TD/TR--PC Motion Encoder, F5, which uses both TD/TR and PC observation features, achieves the highest $\rho$ (0.199). In contrast, F6, which uses only PC observation features, reduces MPJPE by 1.10\% and improves PCK$_{150}$ by 0.916 percentage points compared with F5. These results suggest that, when TD/TR and PC features are already fused during initial pose estimation, using only spatial observation information from PC during pose refinement may be more advantageous in terms of joint position accuracy. For F4 and F5, TD/TR and PC observation tokens are concatenated along the token dimension before observation cross-attention.

\subsection{Discussion}
\label{subsec:experimental_discussion}
The component ablation results show that RME, which learns joint-wise motion representations, contributes the most to performance improvement, while OG, which combines initial pose estimation and observation-guided refinement, provides additional improvements. KME, which performs GT-based motion feature alignment, guides joint-wise feature learning, but its relative improvement remains limited. This suggests that more effective methods for aligning the representation differences between radar observations and GT-derived motion features may be necessary.

On RaMoS-Pose, the PC-only configuration achieves lower MPJPE than the TD/TR-only configuration, while TD/TR--PC fusion provides only limited additional improvements over PC-only. These results suggest that PC provides important information for joint-wise feature learning under the current acquisition conditions, while temporal features from TD/TR may play a complementary role. However, increases in PCK$_{150}$ and $\rho$ resulting from fusion alone are insufficient to conclude that the accuracy of subtle joint movements has improved.

One possible reason for the limited additional improvement is the difference in temporal resolution between radar observations and GT poses. Radar can observe radial movements occurring over short time intervals, whereas some subtle movements may not be sufficiently reflected in GT with lower temporal resolution. In addition, unobserved intervals caused by frame-based radar acquisition may affect temporal feature learning. To investigate these possibilities, future work should employ more continuous radar acquisition and higher-temporal-resolution GT, along with further performance analysis across motion categories and individual joints.


\section{Conclusion}
\label{sec:conclusion}

In this study, we proposed I2R-Pose, which combines motion-aware initial pose estimation with Observation-Guided Refinement. Experimental results showed that joint-wise motion representation learning contributed the most to performance improvement, while observation-guided pose refinement and GT-based motion feature alignment provided additional improvements. On MVDoppler-Pose, I2R-Pose reduced MPJPE by 39.50\% compared with the best-performing existing method. We also constructed RaMoS-Pose, which provides synchronized radar observations and 3D joint GT for diverse indoor motions. On RaMoS-Pose, I2R-Pose reduced MPJPE by 14.42\% with PC-only inputs and by 15.13\% with TD/TR--PC fusion, compared with the best-performing existing PC-based method. These results demonstrate the potential of joint-wise feature learning and sequential pose refinement across different radar representations. Future improvements in motion feature learning are expected to further enhance pose estimation accuracy across diverse pose transitions. Furthermore, the proposed framework could support non-contact human movement monitoring as a radar-based sensing component in IoT-enabled smart homes.



\appendices
\section{Keypoint Motion Descriptor Construction}
\label{app:keypoint_descriptor}

\begin{table*}[!t]
\centering
\caption{Detailed feature composition and mathematical definitions of the 52-dimensional Keypoint Motion Descriptor.}
\label{tab:motion_features_full}

\IIRTableStyle
\renewcommand{\arraystretch}{1.24}

\begin{tabular}{@{}llcl@{}}
\toprule
Feature & Mathematical definition & Dim. & Description \\
\midrule

\multicolumn{4}{@{}l}{
\textbf{Body Geometry (6D): Body geometric structure}
} \\

Root-relative position &
$\mathbf p_{t,j}^{\mathrm{rel}}
=\mathbf p_{t,j}-\mathbf p_{t,r}$ &
3 &
Position relative to the root joint \\

Bone vector &
$\mathbf b_{t,j}
=\mathbf p_{t,j}-\mathbf p_{t,\pi(j)}$ &
3 &
Position relative to the parent joint \\

\addlinespace[3pt]

\multicolumn{4}{@{}l}{
\textbf{Temporal Changes (18D): Temporal changes}
} \\

Backward difference &
$\mathbf d^-_{t,j}
=\mathbf p_{t,j}-\mathbf p_{t-1,j}$ &
3 &
Position change from the previous time step \\

Forward difference &
$\mathbf d^+_{t,j}
=\mathbf p_{t+1,j}-\mathbf p_{t,j}$ &
3 &
Position change to the next time step \\

Centered difference &
$\mathbf d^{\mathrm{c}}_{t,j}
=\frac{\mathbf p_{t+1,j}-\mathbf p_{t-1,j}}{2}$ &
3 &
Centered difference using adjacent time steps \\

Second difference &
$\mathbf d^{(2)}_{t,j}
=\mathbf p_{t+1,j}
-2\mathbf p_{t,j}
+\mathbf p_{t-1,j}$ &
3 &
Second-order temporal difference of positions \\

Bone difference &
$\mathbf d^{\mathrm{bone}}_{t,j}
=\mathbf b_{t,j}-\mathbf b_{t-1,j}$ &
3 &
Temporal change in the bone vector \\

Parent-relative difference &
$\mathbf d^{\mathrm{par}}_{t,j}
=\mathbf d^{\mathrm{c}}_{t,j}
-\mathbf d^{\mathrm{c}}_{t,\pi(j)}$ &
3 &
Movement relative to the parent joint \\

\addlinespace[3pt]

\multicolumn{4}{@{}l}{
\textbf{Motion Statistics (4D): Motion statistics}
} \\

Motion magnitude &
$u_{t,j}
=\sqrt{
\max(
\|\mathbf d^{\mathrm{c}}_{t,j}\|_2^2,
\varepsilon)
}$ &
1 &
Magnitude of the centered difference \\

Squared motion magnitude &
$w_{t,j}
=\|\mathbf d^{\mathrm{c}}_{t,j}\|_2^2$ &
1 &
Squared magnitude of the centered difference \\

Reference-axis difference &
$d^{\mathrm{ref}}_{t,j}
=\mathbf e_{\mathrm{ref}}^\top
\mathbf d^{\mathrm{c}}_{t,j}$ &
1 &
Position change along the reference axis \\

Reference-axis second difference &
$d^{(2),\mathrm{ref}}_{t,j}
=\mathbf e_{\mathrm{ref}}^\top
\mathbf d^{(2)}_{t,j}$ &
1 &
Second-order temporal difference along the reference axis \\

\addlinespace[3pt]

\multicolumn{4}{@{}l}{
\textbf{Multiscale Relations (24D): Multiscale temporal relationships}
} \\

Root temporal cross-products &
$\mathbf c^{\mathrm{root}}_{\tau,t,j}
=\mathbf p^{\mathrm{rel}}_{t+\tau,j}
\times
\mathbf p^{\mathrm{rel}}_{t,j},
\;\tau\in\mathcal T_{\mathrm K}$ &
12 &
Changes in root-relative directions across multiple temporal intervals \\

Bone temporal cross-products &
$\mathbf c^{\mathrm{bone}}_{\tau,t,j}
=\mathbf b_{t+\tau,j}
\times
\mathbf b_{t,j},
\;\tau\in\mathcal T_{\mathrm K}$ &
12 &
Changes in bone directions across multiple temporal intervals \\

\midrule

\multicolumn{2}{@{}l}{
\textbf{Total Motion Descriptor dimensions}
} &
\textbf{52} & \\

\bottomrule
\end{tabular}
\end{table*}

The training-only Keypoint Motion Encoder constructs a
52-dimensional joint-wise motion descriptor from
3D GT joint sequences.
The descriptor captures both relative body geometry and
temporal joint movements and consists of four feature groups:
Body Geometry, Temporal Changes, Motion Statistics, and
Multiscale Relations.
The detailed features and mathematical definitions of each
feature group are presented in Table~\ref{tab:motion_features_full}.

The 3D position of joint $j$ at time step $t$ is denoted by
$\mathbf p_{t,j}\in\mathbb R^3$, where $r$ and $\pi(j)$
denote the indices of the root and parent joints, respectively.
The temporal intervals are defined as
$\mathcal T_{\mathrm K}=\{1,2,4,8\}$, and
$\mathbf e_{\mathrm{ref}}$ denotes a fixed unit vector
along the reference axis.
All temporal differences are computed as frame-wise position
changes without dividing by the actual elapsed time. Temporal differences that are undefined at sequence
boundaries are set to zero.
Temporal cross-products are also set to zero when
$t+\tau>T$, and bone-related features of the root joint,
which has no parent joint, are set to zero.
For numerical stability, $\varepsilon=10^{-6}$ is used,
resulting in a motion magnitude of $u_{t,j}=10^{-3}$
even when the centered difference is zero.

\section{Training Loss Implementation}
\label{app:training_loss}

This section describes the detailed computation of the
pose supervision and motion feature alignment losses
defined in Section~\ref{subsec:training_objectives}.

\subsection{Pose Supervision}

The MPJPE term in Eq.~\eqref{eq:pose_loss} is computed by
averaging the 3D Euclidean distances between predicted
and GT joint positions across the batch, time steps,
and joints.
\begin{equation}
\mathcal L_{\mathrm{MPJPE}}
=\frac{1}{BTJ}
\sum_{b=1}^{B}\sum_{t=1}^{T}\sum_{j=1}^{J}
\left\|
\widehat{\mathbf p}_{b,t,j}
-\mathbf p^{\mathrm{gt}}_{b,t,j}
\right\|_2,
\label{eq:app_mpjpe_loss}
\end{equation}
where $B$, $T$, and $J$ denote the batch size,
number of output time steps, and number of joints,
respectively.

The motion correlation loss follows the publicly
available implementation of MVDoppler-Pose~\cite{mvdoppler_pose}.
For temporal intervals $\tau\in\{2,4,6,8\}$,
the following motion descriptor is constructed
from joint positions:
\begin{equation}
\mathbf m_{\tau,t,j}
=\mathbf p_{t,j}\times\mathbf p_{t+\tau,j}.
\label{eq:app_motion_descriptor}
\end{equation}
The temporal Pearson correlation coefficient $\rho$
between the descriptors constructed from the predicted
joint sequence $\widehat{\mathbf P}$ and the GT joint
sequence $\mathbf P^{\mathrm{gt}}$ is computed for each
joint and coordinate component.
The motion correlation loss is calculated by averaging
$1-\rho$ across the batch, joints, and coordinate
components, followed by equally averaging the losses
over the four temporal intervals.
The arm joint indices are $\{12,13,15,16\}$,
and the leg joint indices are $\{2,3,5,6\}$,
using zero-based joint indexing.
A numerical stability constant $\varepsilon=10^{-6}$
is applied to the denominator of the Pearson correlation,
and descriptors with zero variance are not separately excluded.

\subsection{Motion Feature Alignment}

The temporal feature alignment loss in
Eq.~\eqref{eq:canonical_alignment_loss}
compares the directions and magnitudes of changes
in radar and keypoint features between adjacent time steps.
First, the temporal feature differences are defined as
$\Delta\mathbf Z_{a,t}
=\mathbf Z_{a,t+1}-\mathbf Z_{a,t}$,
where $a\in\{\mathrm R,\mathrm K\}$.

The magnitude of feature changes is computed using
the root mean square (RMS) across the feature channel
dimension at each time step and joint:
\begin{equation}
\begin{aligned}
r_{\mathrm R}
&=\sqrt{\operatorname{mean}_{C_{\mathrm M}}
[(\Delta\mathbf Z_{\mathrm R})^2]+\varepsilon},\\
r_{\mathrm K}
&=\sqrt{\operatorname{mean}_{C_{\mathrm M}}
[(\Delta\mathbf Z_{\mathrm K})^2]}.
\end{aligned}
\label{eq:app_alignment_rms}
\end{equation}
Here, $\varepsilon=10^{-6}$ is a numerical stability
constant applied to the RMS computation of radar features.

When feature changes are very small, their directions
may be unreliable.
Therefore, the direction loss is computed only at
time steps and joints where the RMS of the GT-derived
keypoint features exceeds $10^{-4}$.
The valid set is defined as
$\mathcal V=\{(b,t,j):r_{\mathrm K}(b,t,j)>10^{-4}\}$,
where $b$, $t$, and $j$ denote the indices of the
sample within the batch, time step, and joint,
respectively.
In contrast, the magnitude loss is applied to all
time steps and joints.

The temporal feature alignment loss is defined as
\begin{equation}
\begin{aligned}
\mathcal L_\Delta={}&
\operatorname{mean}_{\mathcal V}\left[
1-\cos_{C_{\mathrm M}}
(\Delta\mathbf Z_{\mathrm R},
\Delta\mathbf Z_{\mathrm K})\right]\\
&+0.25\,\operatorname{SmoothL1}
(r_{\mathrm R},r_{\mathrm K}).
\end{aligned}
\label{eq:app_temporal_alignment}
\end{equation}
Here, $\cos_{C_{\mathrm M}}$ denotes the cosine similarity
computed across the feature channel dimension.
If $\mathcal V$ is empty, the direction loss is set
to zero, while the SmoothL1 loss is averaged over
all time steps and joints.
The SmoothL1 parameter is set to $\beta=1$. The weight $\lambda_\Delta$ of the temporal feature alignment loss is set to zero for the first 10 epochs, linearly increased from 0 to 0.1 between epochs 10 and 40, and then maintained at 0.1 for the remaining epochs.

\section*{Acknowledgment}
We would like to thank Smart Radar System (SRS) for providing the radar sensor used in this research.

%

\bibliographystyle{IEEEtran}
\bibliography{ref}

@book{radar_specularity,
  title={Radar cross section},
  author={Knott, Eugene F and Schaeffer, John F and Tulley, Michael T},
  year={2004},
  publisher={SciTech Publishing}
}

@article{vcf-pose,
  title={VCF-Pose: Velocity-Guided Coarse-to-Fine Pose Estimation from mmWave Point Clouds},
  author={Liu, Xiaokai and Qiao, Lihao and Ji, Zhiyong and Wang, Yangyang and Wang, Jie},
  journal={IEEE Transactions on Cognitive Communications and Networking},
  year={2026},
  publisher={IEEE}
}

@article{multiscale_hpe,
  title={Multi-Scale Spatial-Temporal Representation Learning for mmWave Radar 3D Human Pose Estimation},
  author={Li, Yaxin and Wang, Jun and Yuan, Changshun and Yuan, Xiaoming and Luo, Yuquan and Liang, Song},
  journal={IEEE Internet of Things Journal},
  year={2026},
  publisher={IEEE}
}

@inproceedings{mvdoppler_pose,
  title={{MVDoppler-Pose}: multi-modal multi-view {mmWave} sensing for long-distance self-occluded human walking pose estimation},
  author={Choi, Jaeho and Hor, Soheil and Yang, Shubo and Arbabian, Amin},
  booktitle={Proceedings of the {IEEE/CVF} Conference on Computer Vision and Pattern Recognition},
  pages={27750--27759},
  year={2025}
}

@inproceedings{resnet,
  title={Deep residual learning for image recognition},
  author={He, Kaiming and Zhang, Xiangyu and Ren, Shaoqing and Sun, Jian},
  booktitle={Proceedings of the {IEEE} Conference on Computer Vision and Pattern Recognition},
  pages={770--778},
  year={2016}
}

@article{tang2024mdpose,
author  = {Chong Tang and Wenda Li and Shelly Vishwakarma and Fangzhan Shi and Simon Julier and Kevin Chetty},
title   = {{MDPose}: Human Skeletal Motion Reconstruction Using {WiFi} Micro-Doppler Signatures},
journal = {IEEE Transactions on Aerospace and Electronic Systems},
volume  = {60},
number  = {1},
pages   = {157--167},
year    = {2024},
doi     = {10.1109/TAES.2023.3256973}
}

@article{zhou2023mdpose,
author  = {Xiaolong Zhou and Tian Jin and Yongpeng Dai and Yongkun Song and Zhifeng Qiu},
title   = {{MD-Pose}: Human Pose Estimation for Single-Channel {UWB} Radar},
journal = {IEEE Transactions on Biometrics, Behavior, and Identity Science},
volume  = {5},
number  = {4},
pages   = {449--463},
year    = {2023},
doi     = {10.1109/TBIOM.2023.3265206}
}

@article{sengupta2020mmpose,
author  = {Arindam Sengupta and Feng Jin and Renyuan Zhang and Siyang Cao},
title   = {mm-Pose: Real-Time Human Skeletal Posture Estimation Using mmWave Radars and {CNNs}},
journal = {IEEE Sensors Journal},
volume  = {20},
number  = {17},
pages   = {10032--10044},
year    = {2020},
doi     = {10.1109/JSEN.2020.2991741}
}

@article{sengupta2023mmposenlp,
author  = {Arindam Sengupta and Siyang Cao},
title   = {mmPose-NLP: A Natural Language Processing Approach to Precise Skeletal Pose Estimation Using mmWave Radars},
journal = {IEEE Transactions on Neural Networks and Learning Systems},
volume  = {34},
number  = {11},
pages   = {8418--8429},
year    = {2023},
doi     = {10.1109/TNNLS.2022.3151101}
}

@article{hu2024mmposefk,
author  = {Shuting Hu and Siyang Cao and Nima Toosizadeh and Jennifer Barton and Melvin G. Hector and Mindy J. Fain},
title   = {mmPose-FK: A Forward Kinematics Approach to Dynamic Skeletal Pose Estimation Using mmWave Radars},
journal = {IEEE Sensors Journal},
volume  = {24},
number  = {5},
pages   = {6469--6481},
year    = {2024},
doi     = {10.1109/JSEN.2023.3348199}
}

@article{wu2025mmhpe,
author  = {Yingxiao Wu and Zhongmin Jiang and Haocheng Ni and Changlin Mao and Zhiyuan Zhou and Wenxiang Wang and Jianping Han},
title   = {mmHPE: Robust Multiscale 3-D Human Pose Estimation Using a Single mmWave Radar},
journal = {IEEE Internet of Things Journal},
volume  = {12},
number  = {1},
pages   = {1032--1046},
year    = {2025}
}

@article{chen2006micro,
  title={Micro-Doppler effect in radar: phenomenon, model, and simulation study},
  author={Chen, Victor C and Li, Fayin and Ho, S-S and Wechsler, Harry},
  journal={IEEE Transactions on Aerospace and electronic systems},
  volume={42},
  number={1},
  pages={2--21},
  year={2006},
  publisher={IEEE}
}

@inproceedings{an2022health_mri,
  author    = {An, Sizhe and Li, Yin and Ogras, Umit},
  title     = {{mRI}: Multi-Modal {3D} Human Pose Estimation Dataset Using {mmWave}, {RGB-D}, and Inertial Sensors},
  booktitle = {Advances in Neural Information Processing Systems},
  volume    = {35},
  pages     = {27414--27426},
  year      = {2022}
}

@article{roshandel2025mmprivpose3d,
  author  = {Roshandel, Nima and Scholz, Constantin and Cao, Hoang-Long and Cao, Hoang-Giang and Amighi, Milan and Firouzipouyaei, Hamed and Burkiewicz, Aleksander and Menet, Sebastien and Ballen-Moreno, Felipe and Sisavath, Dylan Warawout and Imrith, Emil and Paolillo, Antonio and Genoe, Jan and Vanderborght, Bram},
  title   = {{mmPrivPose3D}: A {RaDAR}-Based Approach to Privacy-Compliant Pose Estimation and Gesture Command Recognition in Human-Robot Collaboration},
  journal = {IEEE Sensors Journal},
  volume  = {25},
  number  = {15},
  pages   = {29437--29445},
  year    = {2025},
  doi     = {10.1109/JSEN.2025.3578094}
}

@article{wang2023human,
  author  = {Wang, Shuai and Cao, Dongjiang and Liu, Ruofeng and Jiang, Wenchao and Yao, Tianshun and Lu, Chris Xiaoxuan},
  title   = {Human Parsing with Joint Learning for Dynamic {mmWave} Radar Point Cloud},
  journal = {Proceedings of the ACM on Interactive, Mobile, Wearable and Ubiquitous Technologies},
  volume  = {7},
  number  = {1},
  pages   = {34:1--34:22},
  year    = {2023},
  doi     = {10.1145/3580779}
}

@article{kim2009human_microdoppler,
  author  = {Kim, Youngwook and Ling, Hao},
  title   = {Human Activity Classification Based on Micro-{Doppler} Signatures Using a Support Vector Machine},
  journal = {IEEE Transactions on Geoscience and Remote Sensing},
  volume  = {47},
  number  = {5},
  pages   = {1328--1337},
  year    = {2009},
  doi     = {10.1109/TGRS.2009.2012849}
}

@article{zhou2024mdst,
  author  = {Zhou, Xiaolong and Jin, Tian and Dai, Yongpeng and Song, Yongping and Li, Kemeng and Song, Shaoqiu},
  title   = {{MDST}: {2-D} Human Pose Estimation for {SISO UWB} Radar Based on Micro-{Doppler} Signature via Cascade and Parallel {Swin Transformer}},
  journal = {IEEE Sensors Journal},
  volume  = {24},
  number  = {13},
  pages   = {21730--21749},
  year    = {2024},
  doi     = {10.1109/JSEN.2024.3401861}
}

@inproceedings{pavllo20193d,
  title={3d human pose estimation in video with temporal convolutions and semi-supervised training},
  author={Pavllo, Dario and Feichtenhofer, Christoph and Grangier, David and Auli, Michael},
  booktitle={2019 IEEE/CVF Conference on Computer Vision and Pattern Recognition (CVPR)},
  pages={7745--7754},
  year={2019},
  organization={IEEE}
}

@inproceedings{rempe2021humor,
  title={Humor: 3d human motion model for robust pose estimation},
  author={Rempe, Davis and Birdal, Tolga and Hertzmann, Aaron and Yang, Jimei and Sridhar, Srinath and Guibas, Leonidas J},
  booktitle={2021 IEEE/CVF International Conference on Computer Vision (ICCV)},
  pages={11468--11479},
  year={2021},
  organization={IEEE}
}

@inproceedings{chen2022mmbody,
  title={mmbody benchmark: 3d body reconstruction dataset and analysis for millimeter wave radar},
  author={Chen, Anjun and Wang, Xiangyu and Zhu, Shaohao and Li, Yanxu and Chen, Jiming and Ye, Qi},
  booktitle={Proceedings of the 30th ACM International Conference on Multimedia},
  pages={3501--3510},
  year={2022}
}

@article{yang2023mm,
  title={Mm-fi: Multi-modal non-intrusive 4d human dataset for versatile wireless sensing},
  author={Yang, Jianfei and Huang, He and Zhou, Yunjiao and Chen, Xinyan and Xu, Yuecong and Yuan, Shenghai and Zou, Han and Lu, Chris Xiaoxuan and Xie, Lihua},
  journal={Advances in Neural Information Processing Systems},
  volume={36},
  pages={18756--18768},
  year={2023}
}

@article{cui2023milipoint,
  title={Milipoint: A point cloud dataset for mmwave radar},
  author={Cui, Han and Zhong, Shu and Wu, Jiacheng and Shen, Zichao and Dahnoun, Naim and Zhao, Yiren},
  journal={Advances in neural information processing systems},
  volume={36},
  pages={62713--62726},
  year={2023}
}

@inproceedings{ho2024rt,
  title={Rt-pose: A 4d radar tensor-based 3d human pose estimation and localization benchmark},
  author={Ho, Yuan-Hao and Cheng, Jen-Hao and Kuan, Sheng Yao and Jiang, Zhongyu and Chai, Wenhao and Huang, Hsiang-Wei and Lin, Chih-Lung and Hwang, Jenq-Neng},
  booktitle={European conference on computer vision},
  pages={107--125},
  year={2024},
  organization={Springer}
}

@inproceedings{fan2026m4human,
  title={M4human: A large-scale multimodal mmwave radar benchmark for human mesh reconstruction},
  author={Fan, Junqiao and Zhou, Yunjiao and Yang, Yizhuo and Cui, Xinyuan and Zhang, Jiarui and Xie, Lihua and Yang, Jianfei and Lu, Chris Xiaoxuan and Ding, Fangqiang},
  booktitle={Proceedings of the IEEE/CVF Conference on Computer Vision and Pattern Recognition},
  pages={42836--42846},
  year={2026}
}

@article{nguyen2026multi,
  title={A Multi-Sensor Dataset for 3D Human Pose Estimation Using Imaging Radar},
  author={Nguyen, Trung and Rai, Prashant Kumar and Serbenyuk, Nikolay and Ghabcheloo, Reza and Strokina, Nataliya},
  journal={IEEE Open Journal of Signal Processing},
  year={2026},
  publisher={IEEE}
}

@article{zhou2024radarposelifter,
  author  = {Zhou, Xiaolong and Jin, Tian and Dai, Yongpeng and Song, Yongping and Li, Kemeng},
  title   = {Three-Dimensional Human Pose Estimation from Micro-{Doppler} Signature Based on {SISO UWB} Radar},
  journal = {Remote Sensing},
  volume  = {16},
  number  = {7},
  pages   = {1295},
  year    = {2024},
  doi     = {10.3390/rs16071295}
}

@article{an2021mars,
  title={Mars: mmwave-based assistive rehabilitation system for smart healthcare},
  author={An, Sizhe and Ogras, Umit Y},
  journal={ACM Transactions on Embedded Computing Systems (TECS)},
  volume={20},
  number={5s},
  pages={1--22},
  year={2021},
  publisher={ACM New York, NY}
}

@article{wu2022rfmask,
  title={RFMask: A simple baseline for human silhouette segmentation with radio signals},
  author={Wu, Zhi and Zhang, Dongheng and Xie, Chunyang and Yu, Cong and Chen, Jinbo and Hu, Yang and Chen, Yan},
  journal={IEEE Transactions on Multimedia},
  volume={25},
  pages={4730--4741},
  year={2022},
  publisher={IEEE}
}

@article{engel2025advanced,
  title={Advanced millimeter wave radar-based human pose estimation enabled by a deep learning neural network trained with optical motion capture ground truth data},
  author={Engel, Lukas and Mueller, Jonas and Rendon, Eduardo Javier Feria and Dorschky, Eva and Krauss, Daniel and Ullmann, Ingrid and Eskofier, Bjoern M and Vossiek, Martin},
  journal={IEEE Journal of Microwaves},
  volume={5},
  number={2},
  pages={373--387},
  year={2025},
  publisher={IEEE}
}

@inproceedings{zhao2018rf3dskeletons,
  author    = {Zhao, Mingmin and Tian, Yonglong and Zhao, Hang and Alsheikh, Mohammad Abu and Li, Tianhong and Hristov, Rumen and Kabelac, Zachary and Katabi, Dina and Torralba, Antonio},
  title     = {{RF}-Based {3D} Skeletons},
  booktitle = {Proceedings of the ACM SIGCOMM 2018 Conference},
  pages     = {267--281},
  year      = {2018},
  doi       = {10.1145/3230543.3230579}
}

@article{zhang2023mmwavesurvey,
  author  = {Zhang, Jia and Xi, Rui and He, Yuan and Sun, Yimiao and Guo, Xiuzhen and Wang, Weiguo and Na, Xin and Liu, Yunhao and Shi, Zhenguo and Gu, Tao},
  title   = {A Survey of {mmWave}-Based Human Sensing: Technology, Platforms and Applications},
  journal = {IEEE Communications Surveys \& Tutorials},
  volume  = {25},
  number  = {4},
  pages   = {2052--2087},
  year    = {2023},
  doi     = {10.1109/COMST.2023.3298300}
}

@article{shi2025mmpointattention,
  author  = {Shi, Xintong and Bouazizi, Mondher and Ohtsuki, Tomoaki},
  title   = {{mmPoint-Attention}: A Unified Attention Framework for Human Pose and Activity Recognition From {mmWave} Radar Point Clouds},
  journal = {IEEE Sensors Journal},
  volume  = {25},
  number  = {14},
  pages   = {26794--26805},
  year    = {2025},
  doi     = {10.1109/JSEN.2025.3577772}
}

@inproceedings{xue2021mmmesh,
  author    = {Xue, Hongfei and Ju, Yan and Miao, Chenglin and Wang, Yijiang and Wang, Shiyang and Zhang, Aidong and Su, Lu},
  title     = {{mmMesh}: Towards {3D} Real-Time Dynamic Human Mesh Construction Using Millimeter-Wave},
  booktitle = {Proceedings of the 19th Annual International Conference on Mobile Systems, Applications, and Services},
  pages     = {269--282},
  year      = {2021},
  doi       = {10.1145/3458864.3467679}
}

@article{gao2026stargraph,
  author  = {Gao, Senhao and Zhang, Junqing and Mei, Luoyu and Wang, Shuai and Wang, Xuyu},
  title   = {Exploring Spatial-Temporal Representation via Star Graph for {mmWave} Radar-Based Human Activity Recognition},
  journal = {IEEE Transactions on Mobile Computing},
  volume  = {25},
  number  = {4},
  pages   = {5700--5715},
  year    = {2026},
  doi     = {10.1109/TMC.2025.3634221}
}

@article{qiao2022mdseparation,
  author  = {Qiao, Xingshuai and Amin, Moeness G. and Shan, Tao and Zeng, Zhengxin and Tao, Ran},
  title   = {Human Activity Classification Based on Micro-{Doppler} Signatures Separation},
  journal = {IEEE Transactions on Geoscience and Remote Sensing},
  volume  = {60},
  pages   = {1--14},
  note    = {Art. no. 5105014},
  year    = {2022},
  doi     = {10.1109/TGRS.2021.3105124}
}

@article{ren2018stssm,
  author  = {Ren, Lingyun and Tran, Nghia and Foroughian, Farnaz and Naishadham, Krishna and Piou, Jean E. and Kilic, Ozlem and Fathy, Aly E.},
  title   = {Short-Time State-Space Method for Micro-{Doppler} Identification of Walking Subject Using {UWB} Impulse Doppler Radar},
  journal = {IEEE Transactions on Microwave Theory and Techniques},
  volume  = {66},
  number  = {7},
  pages   = {3521--3534},
  year    = {2018},
  doi     = {10.1109/TMTT.2018.2829523}
}

@article{cao2025occmesh,
  author  = {Cao, Haoran and Yu, Jiadi and Kong, Hao and Chen, Yi-Chao and Zhu, Yanmin and Kong, Linghe and Tang, Feilong},
  title   = {{OccMesh}: Occlusion-Aware Multi-User {3D} Human Mesh Reconstruction Using {mmWave} Signals},
  journal = {Proceedings of the ACM on Interactive, Mobile, Wearable and Ubiquitous Technologies},
  volume  = {9},
  number  = {3},
  pages   = {72:1--72:22},
  year    = {2025},
  doi     = {10.1145/3749514}
}

@article{cao2024virteach,
  author  = {Cao, Zhongping and Mei, Guangyu and Guo, Xuemei and Wang, Guoli},
  title   = {{VirTeach}: {mmWave} Radar Point-Cloud-Based Pose Estimation With Virtual Data as a Teacher},
  journal = {IEEE Internet of Things Journal},
  volume  = {11},
  number  = {10},
  pages   = {17615--17628},
  year    = {2024},
  doi     = {10.1109/JIOT.2024.3359209}
}

@article{zhang2024radarfeatures,
  author  = {Zhang, Tao and Qiao, Xingshuai and Li, Xiuping and Bai, Chao},
  title   = {Radar Feature Analysis of Human Activity Recognition Under Multi-View Scenes},
  journal = {IEEE Sensors Journal},
  volume  = {24},
  number  = {14},
  pages   = {21997--22010},
  year    = {2024},
  doi     = {10.1109/JSEN.2023.3325619}
}

@article{song2026star,
  title={STAR: Scene-and Task-Aware 4D Radar Preprocessing Towards End-to-End Cognitive Radar},
  author={Song, Seung-Hyun and Paek, Dong-Hee and Kong, Seung-Hyun},
  journal={arXiv preprint arXiv:2609.24151},
  year={2026}
}



\end{document}